\documentclass[conference,compsoc]{IEEEtran}
\ifCLASSOPTIONcompsoc
  \usepackage[nocompress]{cite}
\else
  \usepackage{cite}
\fi

\ifCLASSINFOpdf
\else
\fi

\usepackage{tikz}
\usepackage{amsmath}
\usepackage{amsfonts}

\usepackage{algorithm}
\usepackage{algpseudocode}
\usepackage{xcolor}
\usepackage[skip=0pt]{subcaption}
\usepackage{longtable}
\usepackage{indentfirst}
\usepackage{dsfont}
\usepackage{bbm}
\usepackage{hyperref}
\usepackage{url} 
\usepackage{graphicx}
\usepackage{multirow}
\usepackage{pifont}
\usepackage{amsthm}
\usepackage{balance}
\usepackage{booktabs}
\usepackage{makecell}
\usepackage{colortbl}

\newtheorem{theorem}{Theorem}

\newtheorem{definition}{Definition}
\newtheorem{corollary}{Corollary}

\newcommand{\changestart}{\begin{color}{black}}
\newcommand{\changeend}{\end{color}}

\begin{document}
%
\title{Preserving Node-level Privacy in Graph Neural Networks}

\author{
{\rm Zihang Xiang}\\
KAUST
\and
{\rm Tianhao Wang}\\
University of Virginia
\and
{\rm Di Wang}\\
KAUST
} 

\maketitle

\begin{abstract}
Differential privacy (DP) has seen immense applications in learning on tabular, image, and sequential data where instance-level privacy is concerned. In learning on graphs, contrastingly, works on \textit{node-level} privacy are highly sparse. Challenges arise as existing DP protocols hardly apply to the message-passing mechanism in Graph Neural Networks (GNNs).

In this study, we propose a solution that specifically addresses the issue of \textit{node-level} privacy. Our protocol has a key component: a sampling routine called \textit{HeterPoisson}, which employs a specialized node sampling strategy and a series of tailored operations to generate a batch of sub-graphs with desired properties. We give the privacy guarantee for \textit{HeterPoisson} by reducing the problem to a mixture of Gaussians mechanism. In addition, our protocol enables GNN learning with good performance, as demonstrated by experiments on five real-world datasets; compared with existing baselines, our method shows significant advantages, especially in the high privacy regime. In the experiment, we also 1) perform membership inference attacks against our protocol and 2) apply privacy audit techniques to confirm our protocol's privacy integrity. 

In the sequel, we present a study on a seemingly appealing approach \cite{sajadmanesh2023gap} (USENIX'23) that protects \textit{node-level} privacy via differentially private node/instance embeddings. Unfortunately, such work has fundamental privacy flaws, which are identified through a thorough case study. More importantly, we prove an impossibility in achieving both (strong) privacy and (acceptable) utility through private instance embedding. The implication is that such an approach has intrinsic utility barriers when enforcing differential privacy.


\end{abstract}

\section{Introduction}

Modern machine learning/deep learning systems exhibit privacy risks: over-fitting by neural networks results in memorization of the training data \cite{DBLP:conf/iclr/ZhangBHRV17} and leads to unacceptable privacy risk \cite{DBLP:conf/sp/ShokriSSS17}; it is known that only sharing model updates still leads to privacy infringements as the adversary can reconstruct the original training data \cite{DBLP:conf/nips/ZhuLH19_deep_leak}. Among all learning tasks, there has been an increasing focus on learning from non-Euclidean data, particularly from graph data, utilizing the message-passing mechanism or Graph Neural Networks (GNNs). Unfortunately, similar to the privacy issues in learning tasks on tabular, sequential, or image data, privacy breaches are also observed when learning GNNs on graphs. 
Wu et al. \cite{DBLP:conf/sp/0011L0022/linkteller} provided evidence that the presence of an edge could be inferred with access to the trained GNN model. There are also membership inference attacks against GNNs under various threat models \cite{
DBLP:conf/icdm/WuYPY21}.

These privacy issues underscore the importance of applying a formal approach to protect data privacy. Among all privacy-enhancing technologies, Differential Privacy (DP) \cite{DBLP:conf/tcc/DworkMNS06} has emerged as a widely adopted approach. Simply speaking, DP introduces calibrated randomness into output intended to be accessible by the public, as this operation conceals the evidence of participation and encourages users' trust \cite{DBLP:conf/sp/TschantzSD20}. Previously, DP has been effectively and extensively applied to preserve instance-level privacy in learning tasks for image data in convolutional neural networks \cite{abadi2016deep}, sequential data in recurrent neural networks \cite{DBLP:conf/iclr/McMahanRT018_dp_lstm}, and language models \cite{yu2022differentially_dpnlp}. Among various private learning protocols, differentially private stochastic gradient descent (DP-SGD, {\it a.k.a.}, NoisySGD) \cite{abadi2016deep,DBLP:conf/globalsip/SongCS13} is a prime example.

For graph data, the ``instance'' whose privacy needs to be protected can be an \textit{edge} or a \textit{node}. The former is called \textit{edge-level} privacy, and the latter \textit{node-level} privacy. In \textit{edge-level} privacy, DP prevents the adversary from confidently inferring whether an \textit{edge} between some nodes exists. In \textit{node-level} privacy, DP prevents the adversary from confidently inferring whether a node has contributed to training a GNN, {\it i.e.}, membership inference attacks. 
For the \textit{edge-level} privacy, there has been some work where the existences of some \textit{edge} are hidden from being inferred \cite{DBLP:conf/ccs/KolluriBHS22,DBLP:conf/sp/0011L0022/linkteller}. 

However, in contrast to all the above successful applications of DP, the protection of \textit{node-level} privacy in GNNs remains largely unanswered. In fact, the success of existing popular applied privacy solutions does not transfer to \textit{node-level} privacy, and existing work tackling this problem fails to give strong results. This suggests that preserving \textit{node-level} privacy in GNN is non-trivial and more challenging. To have a first intuition on such difficulties, the message-passing mechanism requires a node to iteratively aggregate information from its neighbors, thus emphasizing interconnections (dependencies) between nodes; however, existing privacy protocols almost assume data instances are independent during the learning process, {\it i.e.}, settings are conflicting. 

By referencing the DP-SGD protocol, let's dive deeper into some details. The sensitivity analysis, together with the algorithm to ensure bounded sensitivity, is the crucial element in the privacy analysis. Specifically, in the gradient perturbation method, one can enforce bounded sensitivity even if the loss function has no Lipschitz property \cite{abadi2016deep}. This is done by artificially clipping the $\ell_2$ norm of \textit{per-example}'s gradient. However, those techniques for enforcing bounded sensitivity are not readily applicable to GNNs because of the particular behavior in message-passing. Specifically, it is not evident what the \textit{per-example} counterpart should be in GNNs in the first place. Moreover, there are also issues with privacy accounting. It is well-known that sampling on instances leads to privacy amplification, which benefits privacy. However, the sampling on a graph often requires a node to sample its neighbors, and we have difficulties in calculating the sampling rate for a given node, as nodes' edges can be arbitrary. 

The above negative results suggest fundamentally new treatments are needed for preserving \textit{node-level} privacy in GNNs. However, as techniques that can be leveraged are sparse, new challenges arise not only in algorithm design but also in privacy accounting. Recent work demonstrates real-world node privacy risks \cite{DBLP:journals/corr/abs-2102-05429_node_mem}, indicating \textit{node-level} privacy protection is of equal urgency. This is our primary motivation.

\vspace{0.1cm}
\noindent\textbf{This work}. Centered around preserving \textit{node-level} privacy in learning GNNs on graphs, we begin by formulating the privacy problem. We also briefly discuss recent work on this problem and show its limitations. Subsequently, to better understand our design, we present a motivating experiment showing how we trace one node's impact. We find that the more out-edges (pointing to other nodes) a node has, the greater that node impacts the GNN model update. Intuitively, it is because a node has more ``channels'' to deliver impact. Based on such key observations, we form a neighborhood sampling strategy as a countermeasure to offset such impact, which is a part of our method, namely \textit{HeterPoisson}. 

In addition, \textit{HeterPoisson} also enforces independent node sampling behavior to form sub-graphs, and independent trial enables clear and precise privacy analysis. In our algorithmic solution, we take the sub-graph as the \textit{per-example} counterpart to compute the per-sub-graph gradient, followed by enforcing the per-sub-graph gradient with bounded $\ell_2$ norm. Such gradient clipping operation leverages the idea of DP-SGD, {\it i.e.}, separate-then-bound. However, current state-of-the-art privacy accounting methods \cite{sgm} for DP-SGD (Gaussian mechanism with sub-sampling) can't be applied because the settings do not fit. Fortunately, due to the design of \textit{HeterPoisson}, we have a non-trivial privacy guarantee.

In the privacy accounting section, we prove a privacy guarantee due to \textit{HeterPoisson}. We leverage a privacy accounting technique called the mixture of Gaussians mechanism \cite{choquette2023privacy}, which is a generalization of the Gaussian mechanism. The key idea is to treat the sub-graph sampling as a mixture of Gaussians, where each Gaussian corresponds to a sub-graph sampled from a node. Technically, we use privacy loss random variable (PLD) for the mixture of Gaussians mechanism \cite{choquette2023privacy} to find the worst-case and give privacy guarantee for that. Specifically, for each iteration, the privacy loss is dominated by a mixture of Gaussians, and the total privacy loss over $T$ iterations is handled by R\'enyi differential privacy (RDP). Composition by RDP is much easier and to be handled in RDP than that of PLD.

In the evaluation section, our experiments on five real-world graph datasets also demonstrate that our solution provides non-trivial utility; compared with several existing baselines, our performance shows significant advantages, especially in high privacy regimes. Ablation studies are also provided to understand the effectiveness of our design choices. In addition to the performance evaluations, to show our solution's privacy integrity, we include 1) privacy attack experiments to show our solution's resistance to attacks by a strong privacy adversary and 2) privacy audit experiments to show that our privacy accounting has no bugs.

In the final part of this work, we present an in-depth study on a seemingly appealing approach \cite{sajadmanesh2023gap} (USENIX'23) that protects \textit{node-level} privacy via differentially private node/instance embeddings. However, we identify fundamental privacy flaws carried by such work through a case study. More importantly, we show an impossibility result of achieving
both (strong) privacy and (acceptable) utility through private instance embedding. Such results have implications: differentially private instance embedding has intrinsic utility barriers. Accordingly, experiments to confirm our results are also provided. Our study highlights pitfalls that should be avoided in privacy applications. To give a concise summary of our contribution: 
\begin{itemize}

    \item We provide an algorithmic solution together with its privacy accounting for preserving \textit{node-level} privacy in learning GNNs on graphs.
    \vspace{0.05cm}
    \item  We show our solution's effectiveness on real-world datasets, and we also experimentally show our method's privacy integrity and resilience by performing privacy audit and privacy attacks to show our protocol.
     \vspace{0.05cm}
    \item We study and identify privacy flaws for private instance embedding used by some previous work. We also prove an impossibility result for such an approach and provide experimental evaluations.
\end{itemize}

\section{Background}

\subsection{Graph Neural Networks}\label{sec:gnn}

\noindent\textbf{Graph}. A graph is a tuple $\mathcal{G}=(\mathcal{V}, \mathcal{E}, \mathbf{X})$, where $\mathcal{V}$ is the node set, $\mathcal{E}$ is the edge set, and $\mathbf{X} \in \mathbb{R}^{|\mathcal{V}| \times d}$ is the feature matrix whose $i$-th row is a $d$-dimensional vector of node $i$. In node classification tasks, we have additional information $\mathbf{Y}$, which stores the labels for each node with $C$ possible classes. We use $(i\rightarrow j) \in \mathcal{E}$ to denote the existence of an edge from node $i$ to $j$. 
For an undirected graph, $(i\rightarrow j) \in \mathcal{E}$ implies $(j\rightarrow i) \in \mathcal{E}$. 
We say $(i\rightarrow j)\in\mathcal{E}$ is an out-edge of $i$, and an in-edge of $j$. If $(i\rightarrow j)\in \mathcal{E}$, we use $i \in \mathcal{NB}(j)$ to denote $i$ is a neighbor of $j$. 
For node $i$, we use \textit{in-degree} and \textit{out-degree} to denote the size of its in-edges and out-edges, respectively. 

\begin{definition}[GNN]\label{def:gnn}
A graph neural network (GNN) follows the message-passing mechanism: iteratively, it updates each node's embedding by aggregating information from neighbors. The
$k$-th update is formulated as follows. 

\begin{equation}\nonumber
\begin{aligned}
    \mathbf{m}_{\mathcal{NB}(u)}^{(k)}&=\mathbf{AGG}^{(k)}\left(\left\{\mathbf{h}_v^{(k)}| v \in \mathcal{NB}(u)\right\}\right)\\
    \mathbf{h}_u^{(k+1)} & =\phi^{(k)}\left(\mathbf{h}_u^{(k)}, \mathbf{m}_{\mathcal{NB}(u)}^{(k)}\right)
\end{aligned}
\end{equation}
where $\mathbf{h}_u^{(k)}$ is the obtained embedding for node $u$ and $\mathbf{m}_{\mathcal{NB}(u)}^{(k)}$ stands for the aggregated ``message'' from neighbors. $\phi$ is the update function with learnable parameters, often instantiated as a multilayer perceptron (MLP). $\mathbf{AGG}$, which aggregates information, is an arbitrary differentiable function.
\end{definition}

Usually, the final resultant node embedding is used as the representation vector for a node. In GNNs, there are many aggregation methods such as GCN \cite{DBLP:conf/iclr/KipfW17_gcn}, GIN  \cite{DBLP:conf/iclr/XuHLJ19_gin}, SAGE \cite{DBLP:conf/nips/HamiltonYL17_sage}, etc. We present three instantiations of $\mathbf{AGG}$ used in our work in Appendix \ref{appendix:gnns}. We also provide the general training routine for a GNN model in Appendix \ref{appendix:gnns} for reference.

\vspace{0.05cm}
\noindent\textbf{Scenarios}.
There are mainly three kinds of supervised learning task: 1) \textit{node classification} \cite{DBLP:conf/iclr/KipfW17_node_level_task}; 2) \textit{link prediction} \cite{DBLP:conf/esws/SchlichtkrullKB18_edge_level}; 3) \textit{graph classification} \cite{DBLP:conf/nips/YingY0RHL18_graph_level}. Depending on the graph, there are 1) \textit{transductive} setting, where training nodes and testing nodes are on the same graph; 2) \textit{inductive} setting, where training nodes and testing nodes are on different graphs, and the testing nodes are invisible during training.

\subsection{Differential Privacy}

\begin{definition}\label{def:dp}
(Differential Privacy \cite{DBLP:conf/tcc/DworkMNS06})
Given a data universe $\mathcal{X}$, two datasets $X, X'\subseteq \mathcal{X}$ are adjacent if they differ by only one data sample. A randomized algorithm $\mathcal{M}$ is $(\varepsilon,\delta)$-differentially private if for all adjacent datasets $X,\, X'$ and for all events $S$ in the output space of $\mathcal{M}$, we have $\operatorname{Pr}(\mathcal{M}(X)\in S)\leq e^{\varepsilon} \operatorname{Pr}(\mathcal{M}(X')\in S)+\delta$.
\end{definition}

The notion of the adjacent dataset $X,\, X'$ is context-dependent but is often ignored to be discussed. Technically, if $X'$ can be obtained by replacing a data instance of $X$, then it is called \textit{bounded} DP \cite{DBLP:conf/tcc/DworkMNS06}; if $X'$ can be obtained by addition/removal of a data sample of $X$, it is called \textit{unbounded} DP \cite{DBLP:conf/icalp/Dwork06}. Notably, these two notions are equivalent up to a factor of two \cite{DBLP:journals/ftdb/NearH21}. Practically, to have a meaningful privacy guarantee, it is often to set $\varepsilon$ to be some small number, and the $\delta$ can be understood as the failure probability. DP has two notable properties: 1) immune to post-processing; 2) composition: running multiple DP algorithms sequentially also satisfies DP, and the overall privacy parameter can be bounded in terms of individual algorithm's parameters.

\vspace{0.1cm}
\noindent\textbf{Privacy accounting}. Privacy accounting gives the total privacy guarantee for the composition of several (adaptive) private algorithms. We introduce R\'enyi DP (RDP), a relaxation of DP based on  R\'enyi divergence. RDP often serves as a basic analytical tool to analyze the composition of differentially private mechanisms while giving tight results.

\begin{definition}[R\'enyi DP \cite{DBLP:conf/csfw/Mironov17}]\label{def:RDP} The R\'enyi divergence is defined as $\mathcal{D}_{\alpha}(M||N) = \frac{1}{\alpha - 1} \ln \mathbb{E}_{x\sim N}\left[\frac{M(x)}{N(x)}\right]^\alpha$
with $\alpha>1$. 
A randomized mechanism $\mathcal{M}: \mathcal{X}\rightarrow R $ is said to be $(\alpha, \gamma)$-RDP, if 
\begin{equation}\nonumber
\begin{aligned}
    \mathcal{D}_{\alpha}(\mathcal{M}(X)||\mathcal{M}(X')) \leq \gamma
\end{aligned}
\end{equation}
holds for any adjacent dataset $X, X'$.

\end{definition}

\begin{theorem}[Composition by RDP, Theorem 20  \cite{balle2020hypothesis}]\label{thm:rdp_to_dp} For $\alpha>1$ and $\delta>0$, and for a mechanism $\mathcal{M}$ which is $(\alpha, \gamma)$-RDP, the result for $T$-fold adaptive composition of $\mathcal{M}$ satisfies $(\varepsilon, \delta)$-differential privacy and 
$$\varepsilon=T\gamma+\log \frac{\alpha-1}{\alpha}-\frac{\log \delta+\log \alpha}{\alpha-1}.$$
\end{theorem}
Usually, one wants to convert an RDP guarantee to the $(\varepsilon,\delta)$-DP formulation. And Theorem \ref{thm:rdp_to_dp} allows to make such conversion tightly. In practice, the final result is often obtained by optimizing $\varepsilon$ over $\alpha$.

\subsection{Challenges \& Existing Methods }\label{sec:existing_work_des}

\vspace{0.1cm}
\noindent\textbf{Challenges}. Compared to the popular privacy applications in learning on other types of data, solutions for \textit{node-level} privacy in learning on graph data have hardly converged. In the following, we discuss some key challenges.

\textit{1) New problem, few usable techniques.} The effectiveness of existing privacy protocols, including the well-known DP-SGD protocol, does not transfer to GNN training because of the behavior of the message-passing mechanism. At the algorithm level, to bound the sensitivity in DP learning on image, sequential, or tabular data, it suffices to quantify and limit the ``impact'' of \textit{per-example} as data instances are independent. However, in a graph, nodes deliver impact to other nodes upon being aggregated by other nodes, and the counterpart as the \textit{per-example} is ill-defined. In addition to the algorithmic challenge, we also face privacy accounting difficulties. Although current privacy accounting has reached tight results on the sub-sampled Gaussian mechanism \cite{sgm,DBLP:conf/sp/NasrSTPC21} in DP-SGD, unfortunately, it still cannot be applied because it is unknown how to analyze either the sampling rate for a node or the sensitivity. Settings do not fit. Challenges exist both at the algorithm level and privacy accounting level.

\textit{2) How to ensure inference privacy in the transductive setting.} As mentioned before, in the transductive setting, test nodes' aggregation on neighbors can involve training (sensitive) nodes. Hence, the inference on testing nodes possibly leaks sensitive information \cite{DBLP:conf/sp/0011L0022/linkteller}. A trivial countermeasure is to add calibrated noise during inference to ensure certain privacy guarantees. However, this is problematic: 1) adding noise affects the test accuracy; 2) for a fixed privacy budget, only a limited number of inferences/queries are allowed. Thus, this problem also remains unanswered.

\vspace{0.1cm}
\noindent\textbf{Existing methods}. There have been some notable attempts to tackle \textit{node-level} privacy challenges in GNNs. We summarise them in the following.

\textit{Not generalizable}. Olatunji et al. \cite{DBLP:journals/corr/abs-2109-08907} proposed a solution based on the PATE approach \cite{DBLP:conf/iclr/PapernotAEGT17}. However, such an approach only ensures privacy in the prediction process and needs many unlabeled public data for pre-training, making it impractical generally. Daigavane et al. \cite{DBLP:journals/corr/abs-2111-15521-node-level_deva} proposed a method (denoted as NDP) based on \textit{gradient perturbation}. However, limited by the method, the noise added is overwhelming. Specifically, the reported results show that it requires $\varepsilon\geq 15$ (or even $\varepsilon=30$ in some cases) to make utility acceptable. Moreover, referring to the graph settings mentioned in Section \ref{sec:gnn}, NDP fails to consider privacy issues during test/inference under transductive graph setting.

\textit{Incorrect privacy analysis}. Due to the difficulties mentioned above, other works attempt a new type of approach, which is to privately derive the node/instance embeddings for each training node \cite{DBLP:journals/corr/abs-2210-04442_bad_node_level,sajadmanesh2023gap}. The high-level idea is that once all embeddings for each node are derived privately, it is believed that nodes are decoupled from each other. Then, one can leverage existing privacy protocols (such as DP-SGD) to perform downstream private training. 
However, as will be proved in Section \ref{sec:private_node_embedding}, there is a barrier in the utility-privacy trade-off. Specifically, it is impossible for such an approach to have both acceptable utility and strong privacy. We include a detailed case study in Section \ref{sec:case_study} on the analysis by \cite{sajadmanesh2023gap}, which has fundamental privacy analysis flaws.

\section{Privacy Problem Formulation}\label{sec:problems}

In this study, we adopt the \textit{unbounded} DP notion following previous work \cite{DBLP:journals/corr/abs-2111-15521-node-level_deva, sajadmanesh2023gap,abadi2016deep}. As pointed out in Section \ref{sec:existing_work_des} (challenges), the interdependency between nodes due to message-passing is what differentiates \textit{node-level} privacy problem from previous well-studied privacy problems. In the following, we discuss this issue.

\vspace{0.1cm}
\noindent\textbf{Data instances are independent in image datasets}. It is easy to find an interpretation/implication of DP for image datasets in a classification task: whether Alex chooses to contribute his photo or not, the final classifier will not be affected much. Note that the action taken by Alex never modifies other data instances in the dataset. This example represents the independent-data-instance assumption, which is widely used by previous work on privacy \cite{kifer2011no,li2013membership,chen2014correlated}.

\vspace{0.1cm}
\noindent\textbf{Nodes are correlated in graph data}. However, it requires special care when dealing with graph datasets as nodes are connected. Depending on the scenarios, the action of ``with'' or ``without'' the differing node may modify data held by other nodes, and previous works have ignored this discussion. 

To better understand our analysis, we reform the data into tabular form, {\it i.e.}, each row contains the information: $(\textit{a user's node vector}, \textit{a user's out-edges})$. In a real-world application where Twitter (thus, out-edge information is who he/she follows) intends to learn a GNN model on the Twitter network to serve some recommendation purposes \cite{DBLP:conf/kdd/YingHCEHL18_gnn_recomm}, in terms of what DP guarantees, there can be two typical scenarios shown in the following.

\begin{itemize}
\setlength{\itemindent}{-1em}
    \item \textbf{S1}: In this scenario, DP ensures that \textit{whether a new user registers on Twitter or an existing user deletes/erases his/her account}, the final output model will not change much. Intuitively, this leads to that: \textit{the private output makes it hard to infer whether a person is a Twitter user or not}. Consequently, registering or deleting modifies the data stored in other rows (altering the other rows' out-edges information), violating the independent-data-instance assumption. Technically, by Definition \ref{def:dp} of DP, this setting is equivalent to that the data universe $\mathcal{X}$ is all existing and incoming users.
    
    \item \textbf{S2}: Another guarantee is that \textit{whether an existing user chooses to participate in the model training or not}, the final output model will not change much. Intuitively, this leads to that: \textit{the private output makes it hard to infer whether a Twitter user ever participated in the study}.
    Consequently, even though Alex's row is used or never used, other users' data stays intact, as he has no right to force other users to follow or unfollow him. Technically, by Definition \ref{def:dp} of DP, this setting is equivalent to that the data universe $\mathcal{X}$ is all existing users.
\end{itemize}

Apparently, \textbf{S1} and \textbf{S2} represent different scenarios with different types of privacy guarantees. However, \textbf{S1} has some practical issues: 1) it has been argued by Kifer et al. \cite{kifer2011no} that it is not possible to have both privacy and utility if there is no assumption about the interdependency between data instances; moreover, a setting essentially analogous to \textbf{S1} is shown to provide poor utility-privacy trade-offs. 2) Based on the observation that adding/removing a row modifies data in multiple other rows in \textbf{S1}, we may also link this privacy guarantee to group privacy, {\it i.e.}, trying to maintain indistinguishability when a group of data instances changes. Note that a user can be followed by all other users, indicating the group size is $|\mathcal{G}|$ in the worst case, which is too ambitious to satisfy. 3) Although with \textbf{S1}'s privacy guarantee, knowing whether a person uses a prevalent social network like Twitter is probably not so informative (possibly not considered as leaking privacy) in some cases. If true, why bother to ensure DP in \textbf{S1}?

In contrast, \textbf{S2} does not violate the independent-data-instance assumption, and more importantly, \textbf{S2} models a more practical real-world privacy application in general. Our goal is to ensure DP in \textbf{S2}, and we define the \textit{node-level} privacy for \textbf{S2} formally in the following.

\begin{definition}[Node-level Differential Privacy, formal statement for \textbf{S2}]\label{def:node_level_privacy} A private algorithm $\mathcal{L}$ is said to be \textit{node-level} $(\varepsilon, \delta)$-DP with $\varepsilon>0$ and $\delta \in (0,1)$ if for any whole graph $\mathcal{G}$ and any pairs of adjacent graph $\mathcal{G^{*}}\subseteq \mathcal{G}, \mathcal{G'}\subseteq \mathcal{G}$ that differ by a node, and for all events $S\subseteq \mathbb{R}^d$, we have:
\begin{equation}\label{equ:node_dp}\nonumber
\begin{aligned}
    \operatorname{Pr}(\mathcal{L}(\mathcal{G}^*)\in S)\leq e^{\varepsilon} \operatorname{Pr}(\mathcal{L}(\mathcal{G'})\in S)+\delta.
\end{aligned}
\end{equation}

\end{definition}

\vspace{0.1cm}
\noindent\textbf{Formal privacy model}. For any graph $\mathcal{G}$, we aim to ensure differential privacy as defined in Definition \ref{def:node_level_privacy}. In other words, we aim to protect the privacy of sensitive nodes (training nodes that have labels), including all of their node feature vector, edge information, and class information. W.o.l.g., suppose the differing node is $z$ and $\mathcal{G}^* \cup \{z\}= \mathcal{G}'$, information including 1) $z$'s feature vector, 2) label, and 3) in-edge $(i\rightarrow z)\in \mathcal{E}$, out-edge $(z\rightarrow i) \in \mathcal{E}, i\neq z$ will never be queried if $\mathcal{L}$ operates on $\mathcal{G}^*$. And a DP algorithm $\mathcal{L}$ must ensure the distributions of $\mathcal{L}(\mathcal{G}^*)$ and $\mathcal{L}(\mathcal{G}')$ are close. 

Based on such, we can see that the privacy of the node's information we protect is just as strong as that in \textbf{S1}. Following how we model the targeted real-world privacy application, note that even though Alex's information is used or never used, other users' following information will not be modified, as he has no right to force other users to follow or unfollow him when he participates or does not participate. Formally, this means that: in the whole graph $\mathcal{G}$, $\forall$ node $i$, the information of $i$'s in/out-degree stays unchanged upon Alex's choice as the whole graph $\mathcal{G}$ is fixed.

When social network users hesitate to participate in a study $\mathcal{L}$ that adopts GNN models, enforcing  $\mathcal{L}$ is differentially private will significantly boost their trust and encourage participation. We focus on \textit{node classification} GNN task, following previous work \cite{DBLP:journals/corr/abs-2111-15521-node-level_deva, sajadmanesh2023gap}. We consider both inductive and transductive graph settings. We are targeting directed graphs, as it is more general if considering the connectivity (an undirected graph can be treated/processed as directed).

\section{Our \textit{Node-level} Privacy Solution}\label{sec:node_level_solution}

In this section, we first present the experiments that motivate the design of our solution. Then, we present our algorithmic solution with its privacy accounting.

\subsection{Experiments Motivating Our Design}\label{sec:motivation_exp}

\begin{figure}[!ht] 
    \centering
    \includegraphics[width=1\linewidth]{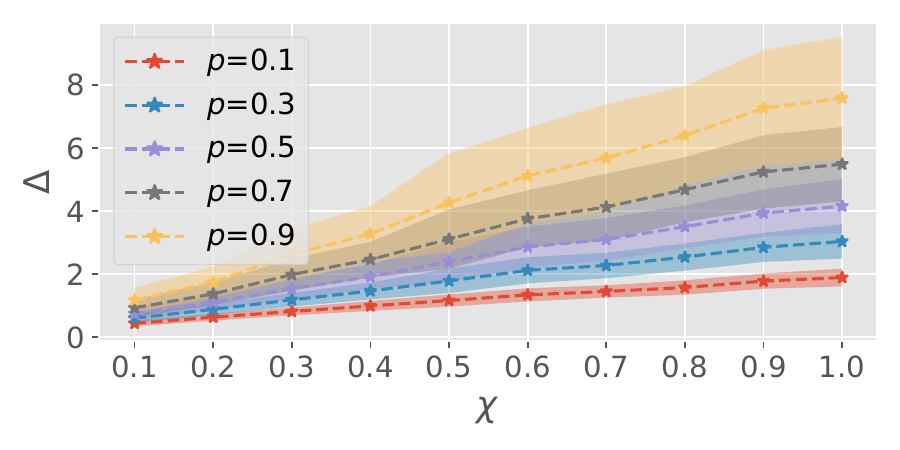}
    
    \caption{Empirical measurements on node $i$'s impact. Parameters are instantiated as $C=10$, $n=100$. The experiment runs 100 times with different seeds.}
    \label{fig:influence} 
    
\end{figure}

\noindent\textbf{Tracing the source: impact measurements}. 
Unlike other data forms, it is not straightforward to see how a node impacts the GNN model update. Nevertheless, we can still get some clues by analyzing the message-passing behavior: node $i$'s information is also propagated to node $j$ if $(i\rightarrow j)\in \mathcal{E}$. Each node $i$ influences other nodes by its out-edges, and this leads to our hypothesis that {\it the greater $|\{(i\rightarrow j)\in \mathcal{E}\}|$ is, the more impact node $i$ has on the GNN model update.}  In what follows, we conduct experiments to measure node $i$'s impact and verify our hypothesis empirically:
\vspace{-0.05cm}
\begin{enumerate}
    \item Initialize a base graph $\mathcal{G}^*$ with $n$ nodes; each node's feature vector is $d$-dimension and is sampled from the standard normal distribution; each node's label is uniformly sampled from $\{1,2,\cdots, C\}$; connecting edges using the \text {Erdős-R\'enyi} model \cite{DBLP:books/cu/Bollobas11_rand_graph} (simply speaking, it connects edges between each pair of nodes independently with some probability $p$).
    \item Initialize a GNN model $w$ and 
    compute the GNN model's gradient $g^*=\nabla f(\mathcal{G}^*;w)$.
    \item Create a new node $i\notin \mathcal{G}^*$ by initializing its feature vector and class in the same way as before; forming $\chi \cdot n$ (where $\chi$ is a parameter we vary) out-edges from $i$ to different random nodes in $\mathcal{G^*}$; denote the resultant graph as $\mathcal{G}'$; compute the GNN model's gradient $g'=\nabla f(\mathcal{G'};w)$. 
    \item Record $\Delta=\|g^*-g'\|_2$. Note that $g^*$ and $g'$ are the results corresponding to the pair of adjacent graphs. Finally, $\Delta$ serves as the proxy for node $i$'s impact on the model update. 
\end{enumerate}

\textbf{Experimental results}. The experimental results are provided in Figure \ref{fig:influence}. We can see that, as expected, when $\chi$ increases (more out-edge $i$ has), $\Delta$ becomes larger (node $i$ has more impact). This can be intuitively explained: the more out-edge $i$ has, the more ``channels'' through which the impact is delivered.

\vspace{0.05cm}
\noindent\textbf{Lessons learned}. Translating to technical terms, by definition, one node's impact on the output in maximum is exactly the ($\ell_2$-)sensitivity of the query function. As shown in \cite{abadi2016deep, DBLP:conf/csfw/Mironov17}, the calibrated DP noise is proportional to $\ell_2$ sensitivity, and the final utility one can get is clearly inverse-related to how much DP noise is added. Hence, to improve the utility, we argue limiting one node's impact or preventing excessive impact on the output is a basic and natural countermeasure. And this leads to our sampling strategy, as will be shown.

\vspace{0.05cm}
\noindent\textbf{Our finding explains why previous works make a particular assumption}. Knowing how one instance affects the query function's output leads to clues to control the sensitivity. In our observations, one node's impact is roughly proportional to its out-edges. This coincides with previous work on \textit{node-level} privacy in GNNs that they assume nodes' degree is bounded by some quantity $\mathrm{D}$, which is assumed to be \textit{much smaller} than the theoretical maximum degree of the graphs for utility reasons \cite{DBLP:journals/corr/abs-2111-15521-node-level_deva, sajadmanesh2023gap}. Using our observation to explain such: bounded degree assumption leads to bounded node's impact. To ensure the bounded degree assumption, previous work \cite{DBLP:journals/corr/abs-2111-15521-node-level_deva, sajadmanesh2023gap} adopts graph projection to erase some edges between nodes. However, we argue that trimming the graph then becomes part of their algorithms, and to avoid privacy leakage, this operation should be treated with extra caution because it is sensitive to one node's presence.

\vspace{0.05cm}
\noindent\textbf{Our design}. In our work, contrastingly, there is no assumption of bounded degree, which sidesteps the above issues. Based on the above experiments, we form a new idea to regulate one node's impact: if one node has larger out-degrees, we will lower the probability for that node to be sampled, offsetting the node's original impact.

\begin{algorithm}[!ht]
\caption{Node-Level DP GNN
}\label{alg:nodedp_training}
\begin{algorithmic}[1]
\small
\renewcommand{\algorithmicrequire}{\textbf{Input:}}
\renewcommand{\algorithmicensure}{\textbf{Output:}}

\Require {Whole graph $\mathcal{G}$, input graph $\hat{\mathcal{G}}\subseteq\mathcal{G}$, initial model $w^0$, number of iteration  $T$, learning rate $\eta$, noise s.t.d. $\sigma$, base sampling rate $q_b$, multiplier $M$, loss function $f(;)$ }

\For{$t=1,2,\cdots,T$}\label{alg:nodedp_training_for}
    \State $G \leftarrow \underline{\mathbf{HeterPoisson}}(\mathcal{G},\hat{\mathcal{G}},q_b,M)$\Comment{Algorithm \ref{alg:HeterPoisson}}\label{alg:nodedp_training_subsample}
    \For{$G_i \in G$} \textbf{in  parallel}\label{alg:nodedp_training_summation_s}
        \State $g_i \gets \nabla f(G_i;w^{t-1})\label{alg:nodedp_training_per_gradient}$\Comment{Per-sub-graph gradient}
        \State  $\hat{g}_i \gets g_i \cdot \mathbf{min}\{1, \frac{1}{2\|g_i\|}\} $ \label{alg:clipping}\Comment{Clip with Threshold 0.5}
    \EndFor\label{alg:nodedp_training_summation_e}
    \State $\Bar{g}^t \gets \sum \hat{g}_i$ \Comment{Non-private output}\label{alg:nodedp_non_private_gradient}
    \State $g^t \gets \Bar{g}^t +\mathcal{N}(0,\sigma\mathbb{I}^d)$ \Comment{Private output}\label{alg:nodedp_private_gradient}
     \State $w ^t \gets w ^{t-1} - \eta g^t$ \Comment{Model update, post-processing}

\EndFor
\Ensure learned model $w^T$
\end{algorithmic}
\end{algorithm}

\begin{algorithm}[!ht]
 \caption{HeterPoisson($\mathcal{G},\hat{\mathcal{G}},q_b,M$)}\label{alg:HeterPoisson}
\begin{algorithmic}[1]
\small
\renewcommand{\algorithmicrequire}{\textbf{Input:}}
\renewcommand{\algorithmicensure}{\textbf{Output:}}

\Require{Whole graph $\mathcal{G}$, input graph $\hat{\mathcal{G}}\subseteq\mathcal{G}$, base sampling rate $q_b$, multiplier $M$}
    \State $G \leftarrow \emptyset$ \Comment{Initialize sub-graph batch container}
    \For{each node $i$ in $\hat{\mathcal{G}}$}\label{alg:HeterPoisson_each_central_s}
    \State $\triangleright$ Forming each sub-graph $G_i$ independently
    \State $p \leftarrow \operatorname{Uniform}(0,1)$
    \If{$p < q_b$}
    \State $\triangleright$ Sampling neighbors of node $i$
    \State $N_i \gets \underline{\mathbf{NeighborSampling}}(\mathcal{G},\mathcal{NB}(i), M)$\Comment{Alg. \ref{alg:poisson_sampling}}\label{alg:HeterPoisson_neighbor_poisson_sampling}
    \State Mark $i$ as \textit{central} node and $N_i$ as \textit{peripheral} nodes\label{alg:sampling_central_node}
    \State Forming the induced sub-graph $G_i$ using $i$ and $N_i$
    \State $G.add(G_i)$  \Comment{Add this sub-graph to the container}\label{alg:HeterPoisson_each_central_sub-graph_sampling}
    
    \EndIf
    \EndFor\label{alg:HeterPoisson_each_central_e}

    \State $\triangleright$ Ensuring no \textit{peripheral} node can be a \textit{central} node \label{alg:HeterPoisson_ensure_no_overlapping}
    \State $\Delta \gets $ all \textit{central} nodes sampled
    \For{$G_i$ in $G$}\label{alg:HeterPoisson_overlapping_s}
        \For{$j$ in \textbf{\textit{peripheral} nodes} of $G_i$}
            \If{$j$ is in $\Delta$}
                \State $\triangleright$ modify this node to be \textbf{NULL}
                \State modify the feature vector of $j$ to be zero in $G_i$
            \EndIf
        \EndFor
    \EndFor\label{alg:HeterPoisson_overlapping_e}
\Ensure{ sub-graph batch $G$}
\end{algorithmic}
\end{algorithm}

\begin{algorithm}[!ht]
\caption{NeighborSampling($\mathcal{G},I, M$)
}\label{alg:poisson_sampling}
\begin{algorithmic}[1]
\small
\renewcommand{\algorithmicrequire}{\textbf{Input:}}
\renewcommand{\algorithmicensure}{\textbf{Output:}}

\Require {Whole graph $\mathcal{G}$, node set $I$, multiplier $M$}

\State $I_s\gets \emptyset$
\For{$i$ in $I$} 
\State $p \leftarrow \operatorname{Uniform}(0,1)$
\State $D_{ot}^i \gets $ $i$'s out-degree in $\mathcal{G}$
\If{$p< M/D_{ot}^i$}\label{alg:neighbor_node_sampling}
\State $I_s.add(i)$
\EndIf
\EndFor
\Ensure Sampled index set $I_s$
\end{algorithmic}
\end{algorithm}

\begin{figure}[!ht] 
    \centering
    
    {\includegraphics[width=1\linewidth]{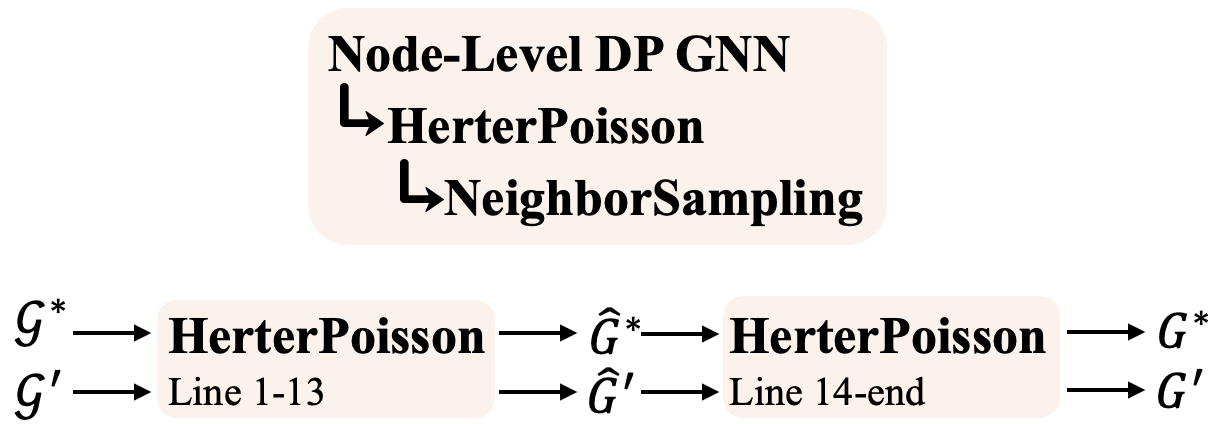}
    \label{fig:illustration_workflow_}}\\
    \caption{Our algorithm structure and the workflow of \textit{HeterPoisson} applied to adjancent graph $\mathcal{G}^*$ and $\mathcal{G}'$. We also highlight the critical stages in \textit{HeterPoisson}.}
    \label{fig:illustration_workflow} 
    
\end{figure}

\subsection{Algorithmic Solution}\label{sec:our_method}

\begin{figure*}[!ht] 
    \centering
    \subfloat[\vspace{0.2cm}Differing node $z$ is sampled as a \textit{central} node]{\includegraphics[width=.9\linewidth]{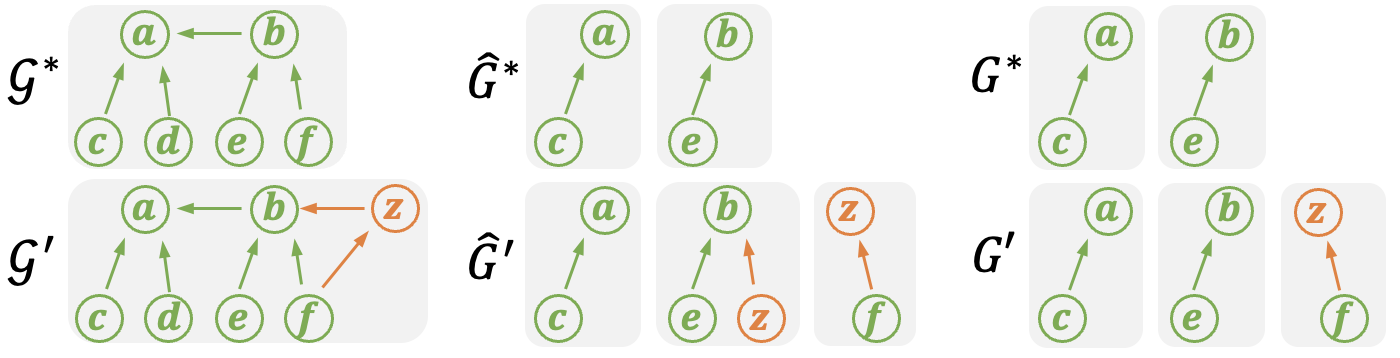}
    \label{fig:illustration_with_z}}\\
    \subfloat[Differing node $z$ is not sampled as a \textit{central} node]{\includegraphics[width=.9\linewidth]{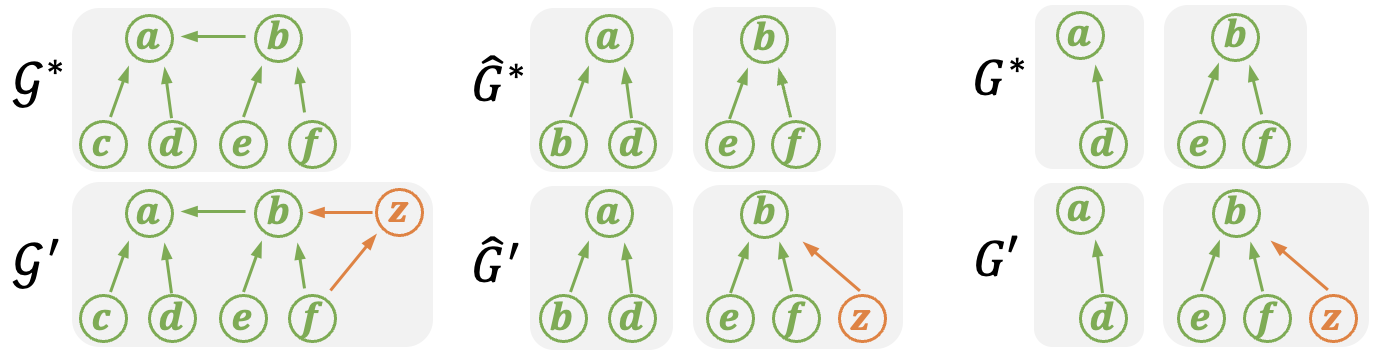}
    \label{fig:illustration_without_z}}\\
    
    \caption{ Critical stages in \textit{HeterPoisson} are highlighted out; we have $\mathcal{G}'=\mathcal{G}^*\cup\{z\}$ where $z$ is the differing node. (a) represents the case when the differing node $z$ is sampled as a \textit{central} node; finally, sub-graph container \textbf{$G'$ has only 1 more sub-graph than  $G^*$}. (b)  represents the case when $z$ is not sampled, $k$ out-pointing nodes of $z$ is sampled (in the above figure, $k=1$), and all of those $k$ nodes also sample $z$ as neighbors; finally, sub-graph container  \textbf{$G^*, G'$ differ in $k$ sub-graphs}, {\it i.e.}, \underline{there are only $k$ sub-graphs in $G^*$ differing from another $k$ sub-graphs in $G'$}.}
    \label{fig:illustration} 
    
\end{figure*}

\noindent\textbf{Algorithm overview}. We present the illustration for the structure of our algorithm in Figure \ref{fig:illustration_workflow}, showing how functions call the others. We also highlight the critical stages in \textit{HeterPoisson} (Algorithm \ref{alg:HeterPoisson}) in Figure \ref{fig:illustration_workflow}. In Figure \ref{fig:illustration}, we show toy examples of critical stages happening in \textit{HeterPoisson} as highlighting those helps better understand our privacy analysis. For either of the adjacent input graphs $\mathcal{G}^*, \mathcal{G}'$,  \textit{HeterPoisson} first receives input graph $\hat{\mathcal{G}}$; then, after many sub-graphs are sampled and formed, we get sub-graph container $\hat{G}$; finally, after those sub-graphs are processed, we get the final sub-graph container $G$. 

\vspace{0.1cm}
\noindent\textbf{Algorithm \ref{alg:nodedp_training}}. The high-level steps of our solution are presented in \textit{}Algorithm \ref{alg:nodedp_training}. In each iteration, \textit{Node-Level DP GNN} first calls function \textit{HeterPoisson} to return a sub-graph container $\hat{G}$ which contains many sub-graphs. This allows us to leverage the ideas from DP-SGD, {\it i.e.}, separate-then-bound; we treat a single sub-graph as the \textit{per-example}, and we clip the per-sub-graph gradient with bounded $\ell_2$-norm. Finally, we add Gaussian noise $\mathcal{N}(0,\sigma^2\mathbb{I}^d)$ \cite{kotz2001laplace} to the sum of clipped gradient vectors to form the private gradient. 


\vspace{0.1cm}
\noindent\textbf{Algorithm \ref{alg:HeterPoisson}}. \textit{HeterPoisson} is the core part of our solution. \textit{HeterPoisson} returns a container that contains many sub-graphs. It first samples the \textit{central} nodes, and based on a \textit{central} node, it calls the function \textit{NeighborSampling} to sample neighbors to form a sub-graph. Note that sampling on \textit{central} nodes are independent trails. After forming many sub-graphs, it ensures no \textit{central} node appears in the \textit{peripheral} nodes in other sub-graphs. We enforce such an operation for utility purposes. Specifically, if we do not have such an operation, although our privacy accounting method can still apply, we need to add more noise as the sensitivity increases due to overlapping nodes among these sub-graphs, which hurts utility. Discussion and experimental results for this study are provided in Section \ref{sec:ablation_study}.

\vspace{0.1cm}
\noindent\textbf{Algorithm \ref{alg:poisson_sampling}}. Similar to the sampling behavior on \textit{central} nodes, \textit{NeighborSampling} also enforces independent trials when sampling on the neighbors of the \textit{central} nodes. For each neighbor, the sampling ratio is adjusted to be inverse-proportional to this neighbor's out-degree in the whole graph $\mathcal{G}$, such that the neighbor's expected number of being sampled as some \textit{peripheral} nodes is $\frac{M}{D_{ot}^i}\times D_{ot}^i=M$. This idea for controlling the impact of nodes comes from our previous motivating experiments in Section \ref{sec:motivation_exp}. On the other hand, we can control $M$ to balance the privacy and utility trade-off. Note that we enforce independent trials, just like when we sample the \textit{central} nodes; this specific operation enables a tractable privacy analysis in Section \ref{sec:privacy_analysis}.

\vspace{0.1cm}
\noindent\textbf{Complexity analysis}. We only elaborate on the additional computational complexities brought by \textit{HeterPoisson}. Notably, in line \ref{alg:HeterPoisson_overlapping_s} to \ref{alg:HeterPoisson_overlapping_e} in Algorithm \ref{alg:HeterPoisson}, the expected running time for the outer ``for'' loop is $\mathcal{O}(q_b|\mathcal{V}|)$, and the expected running time for the inner ``for'' loop is essentially the expected number of \textit{peripheral} sampled, which is $\mathbb{E}_{i\in\mathcal{V}}(\sum_{(j\rightarrow i)\in \mathcal{E}}\frac{M}{D^j_{ot}}) = M$ where $D^j_{ot}$ is out-degree of node $j$. Hence the expected running time in line \ref{alg:HeterPoisson_overlapping_s} to \ref{alg:HeterPoisson_overlapping_e} in Algorithm \ref{alg:HeterPoisson} is $\mathcal{O}(q_b|\mathcal{V}|M)$, which is the additional expected computational complexity brought by Algorithm \ref{alg:HeterPoisson} in each iteration. Note that this is a conservative analysis; in practice, we can parallelize both the outer and inner ``for'' loops, which makes the additional running time almost negligible.

\vspace{0.1cm}
\noindent\textbf{Discussion on \textit{HeterPoisson}}. As suggested by our complexity analysis, our method scales linearly/mildly with the training graph's size ($\mathcal{V}$, number of nodes), making it practical for applications in large graphs. Our method does not assume the specific method of GNN aggregation but only follows the general abstracted aggregation in Definition \ref{def:gnn}. This makes it generalizable to other GNN instances, given alignments with Definition \ref{def:gnn}.  Our method assumes the graph is static, i.e., the graph's nodes/edges are fixed during the training; for other types of graphs, such as dynamic or heterogeneous graphs, our method may not directly apply as the privacy model (e.g., the formulation of adjacent graph dataset pair) is quite different. Substantial additional work may be required to ensure node-level privacy in such graphs.

\changestart

\subsection{Privacy Guarantee}\label{sec:privacy_analysis}

In this section, we prove the privacy guarantee of our algorithm. We need to first introduce the notion of $\varepsilon$-hockey-stick divergence \cite{choquette2023privacy}, which quantifies the privacy loss for some private algorithm.

\begin{definition}[$\varepsilon$-Hockey-Stick Divergence \cite{choquette2023privacy}]\label{def:hockey_stick_divergence}

Given two distributions $P$ and $Q$ over $\mathbb{R}^d$, the $\varepsilon$-hockey-stick divergence is defined as:
\begin{equation}\nonumber
\begin{aligned}
    \operatorname{H}_\varepsilon(P,Q) = \int_x \max \left\{P(x)-e^{\varepsilon} Q(x), 0\right\} \mathrm{d} x.
\end{aligned}
\end{equation}  
\end{definition}
A mechanism $\mathcal{M}$ satisfies $(\varepsilon, \delta)$-DP is equivelent to: for all adjacent databases $D, D^{\prime}$ we have $H_{\varepsilon}\left(\mathcal{M}(D), \mathcal{M}\left(D^{\prime}\right)\right) \leq \delta$. We will also need to introduce the notion of dominance in terms of $\varepsilon$-hockey-stick divergence:
\begin{definition}[$\varepsilon$-Hockey-Stick Dominance \cite{choquette2023privacy}]\label{def:hockey_stick_dominance}
    The privacy loss random variable for $P$ and $Q$ is given by sampling $x \sim P$, and computing $\ln (P(x) / Q(x))$. The PLD of $P$ and $Q$ is the distribution of this random variable. The PLD of $P, Q$ dominates the PLD of $P^{\prime}, Q^{\prime}$ if for any $\varepsilon, H_{\varepsilon}(P, Q) \geq H_{\varepsilon}\left(P^{\prime}, Q^{\prime}\right)$. 
\end{definition}
Intuitively, dominance in terms of PLD means that the privacy loss with respect to $P, Q$ is always larger than that with respect to $P^{\prime}, Q^{\prime}$. We will first introduce a notion the ``mixture of Gaussians (MoG)'' mechanism and then show that our private algorithm reduces to MoG.

\begin{definition}[Mixture of Gauusians (MoG) \cite{choquette2023privacy}]\label{def:moG}
A mixture of Gaussians (MoG) mechanism is defined by two lists, a list of probabilities $\left\{p_1, p_2, \ldots, p_k\right\}$, with $\sum_i p_i=1, p_i \in[0,1]$, a list of sensitivities $\left\{c_1, c_2, \ldots c_k\right\}$ and a noise level $\sigma$. For simplicity, assume $c_i\in\mathbb{R}, c_i \geq 0$. Given dataset $X$, the mechanism $\mathcal{M}_{M o G}\left(\left\{p_1, p_2, \ldots, p_k\right\},\left\{c_1, c_2, \ldots c_k\right\}\right)$ outputs $z \sim N\left(0, \sigma^2\right)$. Given the adjacent dataset $X^{\prime}$, it samples s from the distribution with support $\left\{c_i:{i \in[k]}\right\}$ and associated probabilities $\left\{p_i:{i \in[k]}\right\}$, and outputs $z \sim N\left(s, \sigma^2\right)$. In other words, it is a Gaussian mechanism where the sensitivity $s$ is a random variable distributed according to $\left\{p_i:{i \in[k]}\right\},\left\{c_i:{i \in[k]}\right\}$.
\end{definition}

A counterpart of MoG for $c_i$ being scalars is that $c_i$ can be real vectors $\mathbf{c}_i$. The output is sample from Gaussian $\mathbf{z}\sim\mathcal{N}(\mathbf{0},\sigma^2\mathbb{I})$ or $\mathbf{z}\sim\mathcal{N}(\mathbf{s},\sigma^2\mathbb{I})$ where $\mathbf{s}$ is sample from $\left\{\mathbf{c}_1, \mathbf{c}_2, \ldots \mathbf{c}_k\right\}$ with corresponding probabilities $\left\{p_1, p_2, \ldots, p_k\right\}$. The critical point is that privacy of MoG for real \textbf{vectors} can be reduced to MoG for $c_i$ being \textbf{scalars}:

\begin{corollary}[MoG for real vectors \cite{choquette2023privacy}]\label{cor:vmog_to_mog}
The PLD of
$$
\mathcal{M}_{V M o G}\left(\left\{p_1, p_2, \ldots, p_k\right\},\left\{\mathbf{c}_1, \mathbf{c}_2, \ldots, \mathbf{c}_k\right\}\right)
$$
is dominated by the PLD of
$$
\mathcal{M}_{M o G}\left(\left\{p_1, p_2, \ldots, p_k\right\},\left\{\left\|\mathbf{c}_1\right\|_2,\left\|\mathbf{c}_2\right\|_2, \ldots,\left\|\mathbf{c}_k\right\|_2\right\}\right) .
$$
\end{corollary}
Given all the preparations, we have the following theorem.

\begin{theorem}\label{thm:privacy_guarantee}
    Define distribution $P$ as $\mathcal{N}(0,\sigma^2)$ and $Q$ as $\mathcal{N}(s,\sigma^2)$, where $s$ is a random variable sampled from the distribution with support $$\left\{c_i:i=0,1/2,1,2,3,\cdots,D_{ot}\right\}$$ and associated probabilities $$\left\{p_i:i=0,1/2,1,2,3,\cdots,D_{ot}\right\}$$ for some integer $D_{ot}>0$. And 
    $
    c_i = i,
    $
    \[
    p_i = 
    \begin{cases}
        (1-q_b)\mathbf{BI}(i, D_{ot}, \frac{q_b M}{D_{ot}}) & \text{for } i \neq 1/2 \\
        q_b & \text{for } i = 1/2
    \end{cases}
    \]
    where $\mathbf{BI}(i, D_{ot}, \frac{q_b M}{D_{ot}})$ is the PMF of binomial distribution with parameters $D_{ot}$ and $q_b M/D_{ot}$. The mechanism $\mathcal{M}$ in Algorithm \ref{alg:nodedp_training} satisfies $(\alpha,\gamma)$-RDP where 
    $$
    \gamma\leq T\cdot \max_{D_{ot}\in [|\mathcal{G}|-1]} \max(\mathcal{D}_{\alpha}(P||Q),\mathcal{D}_{\alpha}(Q||P))
    $$
\end{theorem}

\noindent\textbf{Proof sketch}. The proof is based on the fact that the output of Algorithm \ref{alg:nodedp_training} is a mixture of Gaussians (MoG) mechanism, and we can reduce the privacy analysis to the MoG for scalars. The key point is that the sensitivity $s$ is a random variable sampled from the distribution with support $\left\{c_i:i=0,1/2,1,2,3,\cdots,D_{ot}\right\}$ and associated probabilities $\left\{p_i:i=0,1/2,1,2,3,\cdots,D_{ot}\right\}$. First, for any node, the probability of it being sampled as a central node is $q_b$. In this case, the sensitivity is $1/2$ because we enforce that there is no overlapping between the central node and the peripheral nodes. Conditioned on some node not being sampled as a central node, which happens with probability $1-q_b$, the analysis of the sensitivity aligns with a binomial distribution. The detailed proof is provided in Appendix \ref{app:proof_privacy_guarantee}.

There are some technical points that we need to clarify. We use dominance in terms of PLD to find the worst-case privacy loss and use RDP to handle the composition (multiplication by $T$), this is because composition by RDP is much easier to handle than composition by PLD. The price we pay is that this may result in a little bit looser bound, but it is still a good bound that works for our algorithm. 

\changeend

\subsection{Preserve Privacy at Inference}\label{sec:inference_privacy}

Another important functionality of \textit{HeterPoisson} is to ensure inference privacy at test time under the transductive setting as discussed in Section \ref{sec:problems}. We have shown that adding noise during inference is problematic, and to address this issue, we devise a simple yet effective approach. Given all of our previous work, it only requires modifying one line in our algorithm: during test/inference, in line \ref{alg:HeterPoisson_neighbor_poisson_sampling} in Algorithm \ref{alg:HeterPoisson}, instead of passing the whole neighboring nodes of the current test nodes as the function argument, we only pass the neighboring nodes which are not training nodes, {\it i.e.}, test nodes can only sample nodes within the test nodes set, not from the training nodes set. That is, sensitive information is insulated from being accessed at test time, hence no more privacy concerns.

\begin{table}[!ht] 
\scriptsize
    \centering
    \begin{tabular}{c|c|c|c|c|c}
    \toprule
     & \textbf{Facebook} &  \textbf{Twitch} & \textbf{Amazon} & \textbf{PubMed} & \textbf{Reddit} \\
     \midrule
    $\#$ nodes & $22$K  & $9$K   & $13$K & $19$K & $232$K \\
    $\#$ edges & $342$K & $315$K & $491$K & $88$K & $114$M \\
     Avg. $\#$ edges & $15$ & $33$ & $35$ & $5$ & $492$ \\
     \midrule
     Features & $128$ & $128$ & $767$ & $500$ & $602$\\
     Classes  & $4$   & $2$   & $10$   & $3$   & $41$ \\
    \bottomrule
    \end{tabular}
    \caption{Dataset statistics}
    \label{tab:dataset_summary}

\end{table}

\begin{table*}[!ht] 
\begin{subtable}[h]{\textwidth}
    \raggedleft
    \resizebox{\columnwidth}{3.1cm}{
    \begin{tabular}{l*{15}{|c}}
    \toprule
    \multirow{2}{*}{{\Large $\varepsilon$}} & \multicolumn{5}{c|}{GCN} & \multicolumn{5}{c|}{GIN} & \multicolumn{5}{c}{SAGE}\\
    & \textbf{Facebook} & \textbf{Twitch} & \textbf{Amazon} & \textbf{PubMed} & \textbf{Reddit} & \textbf{Facebook} & \textbf{Twitch} & \textbf{Amazon} & \textbf{PubMed} & \textbf{Reddit} & \textbf{Facebook} & \textbf{Twitch} & \textbf{Amazon} & \textbf{PubMed} & \textbf{Reddit} \\
    
\midrule
\multirow{4}{*}{2}&{{\large ${48.6}_{\pm 3.7}$}}&{{\large ${59.0}_{\pm 2.8}$}}&{{\large ${37.4}_{\pm 0.9}$}}&{{\large ${39.3}_{\pm 0.5}$}}&{{\large ${41.9}_{\pm 1.6}$}}&{\large $-$}&{\large $-$}&{\large $-$}&{\large $-$}&{\large $-$}&{\large $-$}&{\large $-$}&{\large $-$}&{\large $-$}&{\large $-$}\\
&{\large $34.3_{\pm 0.9}$}&{\large $55.9_{\pm 0.3}$}&{\large $34.9_{\pm 0.9}$}&{\large $41.0_{\pm 1.2}$}&{\large $25.4_{\pm 0.3}$}&{\large $-$}&{\large $-$}&{\large $-$}&{\large $-$}&{\large $-$}&{\large $-$}&{\large $-$}&{\large $-$}&{\large $-$}&{\large $-$}\\
&{\large $19.1_{\pm 0.1}$}&{\large $13.7_{\pm 0.5}$}&{\large $24.7_{\pm 0.4}$}&{\large $38.8_{\pm 0.3}$}&{\large $39.2_{\pm 0.3}$}&{\large $32.7_{\pm 0.9}$}&{\large $4.9_{\pm 0.9}$}&{\large $4.6_{\pm 0.9}$}&{\large $32.0_{\pm 0.5}$}&{\large $35.3_{\pm 0.3}$}&{\large $19.1_{\pm 0.6}$}&{\large $2.3_{\pm 0.4}$}&{\large $22.1_{\pm 0.9}$}&{\large $39.9_{\pm 0.9}$}&{\large $35.4_{\pm 0.4}$}\\
&\cellcolor{gray!30}\underline{{\large $\textbf{74.2}_{\pm 1.0}$}}&\cellcolor{gray!30}\underline{{\large $\textbf{65.8}_{\pm 0.2}$}}&\cellcolor{gray!30}\underline{{\large $\textbf{77.5}_{\pm 0.4}$}}&\cellcolor{gray!30}\underline{{\large $\textbf{78.0}_{\pm 0.3}$}}&\cellcolor{gray!30}\underline{{\large $\textbf{83.2}_{\pm 0.3}$}}&\cellcolor{gray!30}\underline{{\large $\textbf{74.7}_{\pm 0.8}$}}&\cellcolor{gray!30}\underline{{\large $\textbf{65.3}_{\pm 0.5}$}}&\cellcolor{gray!30}\underline{{\large $\textbf{76.9}_{\pm 0.2}$}}&\cellcolor{gray!30}\underline{{\large $\textbf{82.1}_{\pm 0.4}$}}&\cellcolor{gray!30}\underline{{\large $\textbf{81.1}_{\pm 0.4}$}}&\cellcolor{gray!30}\underline{{\large $\textbf{74.4}_{\pm 0.5}$}}&\cellcolor{gray!30}\underline{{\large $\textbf{65.8}_{\pm 0.4}$}}&\cellcolor{gray!30}\underline{{\large $\textbf{76.7}_{\pm 0.2}$}}&\cellcolor{gray!30}\underline{{\large $\textbf{80.7}_{\pm 0.5}$}}&\cellcolor{gray!30}\underline{{\large $\textbf{81.0}_{\pm 0.4}$}}\\
\midrule
\multirow{4}{*}{4}&{{\large ${51.0}_{\pm 3.6}$}}&{{\large ${60.2}_{\pm 0.8}$}}&{{\large ${37.4}_{\pm 0.9}$}}&{{\large ${39.3}_{\pm 0.5}$}}&{{\large ${42.7}_{\pm 1.5}$}}&{\large $-$}&{\large $-$}&{\large $-$}&{\large $-$}&{\large $-$}&{\large $-$}&{\large $-$}&{\large $-$}&{\large $-$}&{\large $-$}\\
&{\large $35.2_{\pm 0.4}$}&{\large $59.0_{\pm 0.4}$}&{\large $30.9_{\pm 0.6}$}&{\large $34.8_{\pm 0.4}$}&{\large $25.5_{\pm 0.9}$}&{\large $-$}&{\large $-$}&{\large $-$}&{\large $-$}&{\large $-$}&{\large $-$}&{\large $-$}&{\large $-$}&{\large $-$}&{\large $-$}\\
&{\large $48.5_{\pm 0.2}$}&{\large $32.7_{\pm 0.4}$}&{\large $30.5_{\pm 0.9}$}&{\large $37.4_{\pm 0.5}$}&{\large $50.9_{\pm 0.9}$}&{\large $47.2_{\pm 0.4}$}&{\large $32.2_{\pm 0.9}$}&{\large $13.4_{\pm 0.9}$}&{\large $39.2_{\pm 0.4}$}&{\large $45.1_{\pm 0.9}$}&{\large $38.4_{\pm 0.3}$}&{\large $15.1_{\pm 0.5}$}&{\large $22.9_{\pm 0.2}$}&{\large $40.5_{\pm 0.9}$}&{\large $49.2_{\pm 0.2}$}\\
&\cellcolor{gray!30}\underline{{\large $\textbf{74.8}_{\pm 1.1}$}}&\cellcolor{gray!30}\underline{{\large $\textbf{66.4}_{\pm 0.4}$}}&\cellcolor{gray!30}\underline{{\large $\textbf{80.6}_{\pm 0.4}$}}&\cellcolor{gray!30}\underline{{\large $\textbf{78.7}_{\pm 0.1}$}}&\cellcolor{gray!30}\underline{{\large $\textbf{84.9}_{\pm 0.3}$}}&\cellcolor{gray!30}\underline{{\large $\textbf{75.8}_{\pm 0.5}$}}&\cellcolor{gray!30}\underline{{\large $\textbf{66.1}_{\pm 1.0}$}}&\cellcolor{gray!30}\underline{{\large $\textbf{80.6}_{\pm 0.9}$}}&\cellcolor{gray!30}\underline{{\large $\textbf{83.8}_{\pm 0.4}$}}&\cellcolor{gray!30}\underline{{\large $\textbf{83.2}_{\pm 0.2}$}}&\cellcolor{gray!30}\underline{{\large $\textbf{74.9}_{\pm 0.6}$}}&\cellcolor{gray!30}\underline{{\large $\textbf{66.5}_{\pm 0.3}$}}&\cellcolor{gray!30}\underline{{\large $\textbf{79.9}_{\pm 0.4}$}}&\cellcolor{gray!30}\underline{{\large $\textbf{81.7}_{\pm 0.3}$}}&\cellcolor{gray!30}\underline{{\large $\textbf{83.1}_{\pm 0.3}$}}\\
\midrule
\multirow{4}{*}{8}&{{\large ${52.1}_{\pm 3.0}$}}&{{\large ${60.4}_{\pm 0.8}$}}&{{\large ${37.4}_{\pm 0.9}$}}&{{\large ${39.3}_{\pm 0.5}$}}&{{\large ${43.4}_{\pm 1.4}$}}&{\large $-$}&{\large $-$}&{\large $-$}&{\large $-$}&{\large $-$}&{\large $-$}&{\large $-$}&{\large $-$}&{\large $-$}&{\large $-$}\\
&{\large $34.1_{\pm 1.2}$}&{\large $56.7_{\pm 0.9}$}&{\large $36.1_{\pm 0.9}$}&{\large $33.3_{\pm 0.3}$}&{\large $29.1_{\pm 0.4}$}&{\large $-$}&{\large $-$}&{\large $-$}&{\large $-$}&{\large $-$}&{\large $-$}&{\large $-$}&{\large $-$}&{\large $-$}&{\large $-$}\\
&{\large $56.9_{\pm 0.4}$}&{\large $61.6_{\pm 0.2}$}&{\large $27.9_{\pm 0.6}$}&{\large $38.3_{\pm 0.3}$}&{\large $62.2_{\pm 0.6}$}&{\large $64.9_{\pm 0.4}$}&{\large $41.8_{\pm 0.9}$}&{\large $26.1_{\pm 0.9}$}&{\large $47.7_{\pm 0.4}$}&{\large $54.9_{\pm 0.4}$}&{\large $52.7_{\pm 0.3}$}&{\large $56.8_{\pm 0.3}$}&{\large $32.0_{\pm 0.9}$}&{\large $39.3_{\pm 0.4}$}&{\large $62.4_{\pm 0.5}$}\\
&\cellcolor{gray!30}\underline{{\large $\textbf{74.9}_{\pm 1.2}$}}&\cellcolor{gray!30}\underline{{\large $\textbf{66.4}_{\pm 0.3}$}}&\cellcolor{gray!30}\underline{{\large $\textbf{82.6}_{\pm 0.1}$}}&\cellcolor{gray!30}\underline{{\large $\textbf{79.4}_{\pm 0.2}$}}&\cellcolor{gray!30}\underline{{\large $\textbf{85.9}_{\pm 0.3}$}}&\cellcolor{gray!30}\underline{{\large $\textbf{76.6}_{\pm 0.5}$}}&\cellcolor{gray!30}\underline{{\large $\textbf{66.2}_{\pm 0.5}$}}&\cellcolor{gray!30}\underline{{\large $\textbf{83.0}_{\pm 1.2}$}}&\cellcolor{gray!30}\underline{{\large $\textbf{84.7}_{\pm 0.3}$}}&\cellcolor{gray!30}\underline{{\large $\textbf{84.5}_{\pm 0.2}$}}&\cellcolor{gray!30}\underline{{\large $\textbf{75.5}_{\pm 0.7}$}}&\cellcolor{gray!30}\underline{{\large $\textbf{66.7}_{\pm 0.1}$}}&\cellcolor{gray!30}\underline{{\large $\textbf{82.7}_{\pm 0.5}$}}&\cellcolor{gray!30}\underline{{\large $\textbf{82.4}_{\pm 0.0}$}}&\cellcolor{gray!30}\underline{{\large $\textbf{84.4}_{\pm 0.2}$}}\\
\midrule
\multirow{4}{*}{16}&{{\large ${52.9}_{\pm 2.7}$}}&{{\large ${60.4}_{\pm 0.8}$}}&{{\large ${37.4}_{\pm 0.9}$}}&{{\large ${39.3}_{\pm 0.5}$}}&{{\large ${44.0}_{\pm 1.4}$}}&{\large $-$}&{\large $-$}&{\large $-$}&{\large $-$}&{\large $-$}&{\large $-$}&{\large $-$}&{\large $-$}&{\large $-$}&{\large $-$}\\
&{\large $32.6_{\pm 0.1}$}&{\large $56.9_{\pm 0.9}$}&{\large $34.6_{\pm 1.2}$}&{\large $31.4_{\pm 0.4}$}&{\large $34.8_{\pm 1.2}$}&{\large $-$}&{\large $-$}&{\large $-$}&{\large $-$}&{\large $-$}&{\large $-$}&{\large $-$}&{\large $-$}&{\large $-$}&{\large $-$}\\
&{\large $70.3_{\pm 0.1}$}&{\large $63.7_{\pm 0.2}$}&{\large $41.2_{\pm 0.9}$}&{\large $40.3_{\pm 0.9}$}&{\large $67.9_{\pm 0.3}$}&{\large $75.2_{\pm 0.9}$}&{\large $50.5_{\pm 0.9}$}&{\large $38.9_{\pm 0.9}$}&{\large $52.0_{\pm 0.9}$}&{\large $59.9_{\pm 0.9}$}&{\large $67.3_{\pm 0.1}$}&{\large $64.4_{\pm 0.4}$}&{\large $32.8_{\pm 0.3}$}&{\large $39.6_{\pm 0.9}$}&{\large $69.9_{\pm 0.9}$}\\
&\cellcolor{gray!30}\underline{{\large $\textbf{75.2}_{\pm 1.1}$}}&\cellcolor{gray!30}\underline{{\large $\textbf{66.9}_{\pm 0.2}$}}&\cellcolor{gray!30}\underline{{\large $\textbf{83.5}_{\pm 0.3}$}}&\cellcolor{gray!30}\underline{{\large $\textbf{79.4}_{\pm 0.4}$}}&\cellcolor{gray!30}\underline{{\large $\textbf{86.7}_{\pm 0.3}$}}&\cellcolor{gray!30}\underline{{\large $\textbf{76.6}_{\pm 0.8}$}}&\cellcolor{gray!30}\underline{{\large $\textbf{66.5}_{\pm 0.2}$}}&\cellcolor{gray!30}\underline{{\large $\textbf{84.3}_{\pm 1.3}$}}&\cellcolor{gray!30}\underline{{\large $\textbf{85.4}_{\pm 0.3}$}}&\cellcolor{gray!30}\underline{{\large $\textbf{85.6}_{\pm 0.2}$}}&\cellcolor{gray!30}\underline{{\large $\textbf{75.6}_{\pm 0.6}$}}&\cellcolor{gray!30}\underline{{\large $\textbf{67.0}_{\pm 0.3}$}}&\cellcolor{gray!30}\underline{{\large $\textbf{83.5}_{\pm 0.1}$}}&\cellcolor{gray!30}\underline{{\large $\textbf{82.9}_{\pm 0.1}$}}&\cellcolor{gray!30}\underline{{\large $\textbf{85.5}_{\pm 0.2}$}}\\
\bottomrule

\end{tabular}
    }

\end{subtable}

\caption{Classification accuracy. In each cell, from top to bottom, the result is \textit{naive DP-SGD}, GAP \cite{sajadmanesh2023gap}, NDP \cite{DBLP:journals/corr/abs-2111-15521-node-level_deva}, and \underline{\textbf {our method}}. Best results are shaded in gray. As \textit{naive DP-SGD} and GAP are not related to any GNN model, their results are only presented in the ``GCN'' column.}
\label{tab:compare_with_GAP_and_deva}

\end{table*}

\section{Method Evaluation}\label{sec:eval}

\noindent\textbf{Organization}. We first show the performance of our method by comparing it with various approaches; we also provide ablations studies on our design choices. Finally, we 1) apply current privacy audit approaches to test the privacy integrity of our method and 2) perform membership inference attacks by a strong adversary to test our protocol's resilience. 
Our implementation is in an anonymous 
link\footnote{{
https://github.com/zihangxiang/PNPiGNNs.git
}}.  

\noindent\textbf{Datasets}. We test our method on five datasets with various properties: Facebook \cite{rozemberczki2021multi}, Twitch \cite{rozemberczki2021multi}, Reddit \cite{DBLP:conf/nips/HamiltonYL17_sage}, Amazon \cite{shchur2018pitfalls}, and PubMed \cite{yang2016revisiting}. The first three are related to social network applications where node/user privacy is concerned; the other two datasets with different properties are included for completeness. The summary for these datasets is presented in Table  \ref{tab:dataset_summary}. We test our approach under both transductive and inductive settings.


\noindent\textbf{Parameters}. To set up the graph for experiments, the training and testing nodes are randomly split by $(80\%,20\%)$ from the original graph. We set the neighborhood sampling multiplier $M$ to be the one lead to best performance among $\{1,2,3,4\}$.
For privacy parameters, we set $\delta = \frac{1}{|\mathcal{V}|^{1.1}}$ and vary $\varepsilon$ to be $\{2,4,8,16\}$ for all datasets. 
We set $q_b=\frac{4096}{|\mathcal{V}|}$ for all datasets and  $T=\lceil\frac{9}{q_b}\rceil$ for Facebook, Twitch, Amazon and PubMed, $T=\lceil\frac{4}{q_b}\rceil$ for Reddit. We use $N_{test}$ to denote the number of neighbors to samples during testing, and $N_{test}=13$ unless otherwise re-clarified. 

\noindent\textbf{Network setup}. For the neural network setup, we instantiate the update function $\phi$ to be a multilayer perceptron (MLP) with $elu$ non-linear activation; the input dimension is determined by the dimension of the node's feature vectors, and we use the same hidden dimension setup $hid\_dim=128$ for all GNN models.

\noindent\textbf{Software implementation:} Our implementation builds on top of PyTorch Geometric 2.2.0
and Pytorch 1.13. 
To efficiently parallelize per-sub-graph gradient computation, we leverage \textit{functorch}'s \textit{vmap} primitive.

\subsection{Performance Comparison}\label{sec:existing_work}

In Table \ref{tab:compare_with_GAP_and_deva}, we first observe that the \textit{naive DP-SGD}  fails to achieve satisfying utility, although such an approach is straightforward. To re-describe the \textit{naive DP-SGD} approach, it only takes the node feature vector as the input, and it treats one feature vector just as it treats an image in the well-studied DP-SGD applications on image data \cite{abadi2016deep}. These experimental results show that the effectiveness of DP-SGD does not transfer to GNN, and ignoring edge information is sub-optimal. In the following comparison, we aim to show that our solution also has a significant advantage over existing baselines. Results are also presented in Table \ref{tab:compare_with_GAP_and_deva}.

One notable work included for comparison is GAP \cite{sajadmanesh2023gap}.  As discussed in Section \ref{sec:private_node_embedding} and Section \ref{sec:case_study}, GAP significantly underestimated the DP noise needed in its aggregation stage. Therefore, we correct the amount of DP noise in this experiment. We can observe that our solution outperforms GAP across each dataset and each privacy level tested. Another notable work is NDP \cite{DBLP:journals/corr/abs-2111-15521-node-level_deva}. Recall that NDP fails to provide inference privacy mentioned in Section \ref{sec:inference_privacy}, nevertheless, we ignore such weakness of NDP in the experiment. Also recall that in Section \ref{sec:existing_work_des}, we mentioned that NDP ensures DP by adding excessive noise, and it is reported in NDP that 1) it needs $\varepsilon\geq 15$ (or $\varepsilon\approx 30$ in some cases) to have acceptable utility; 2) performance in high privacy regime ({\it e.g.}, $\varepsilon=2$) is trivial (close to random gussing). We can also observe this fact from Table \ref{tab:compare_with_GAP_and_deva} that, when $\varepsilon=2$, the test accuracy of NDP is very low, which is consistent with the conclusions given by \cite{DBLP:journals/corr/abs-2111-15521-node-level_deva}. 

Note that Table \ref{tab:compare_with_GAP_and_deva} shows the results for transductive settings, and the results under inductive settings are provided in Appendix \ref{app:add_exp} (Table \ref{tab:compare_with_GAP_and_deva_inductive}). All results show that our method's performance leads by a large margin compared with existing baselines, especially in the high privacy regime. We also report the results of classification precision in Table \ref{tab:compare_with_GAP_and_deva_precision}, which is relevant to the discussion in Section \ref{sec:private_node_embedding}.

\subsection{Ablation Studies}\label{sec:ablation_study}

In this part, we present the ablation studies on our design 
choices and some other aspects that provide some guidance in the parameter selections in \textit{node-level} privacy-preserving GNN applications.


%
    
%

\begin{figure}[!ht] 
    \centering
    \subfloat[$\varepsilon=2$]
    {\includegraphics[width=.49\linewidth]{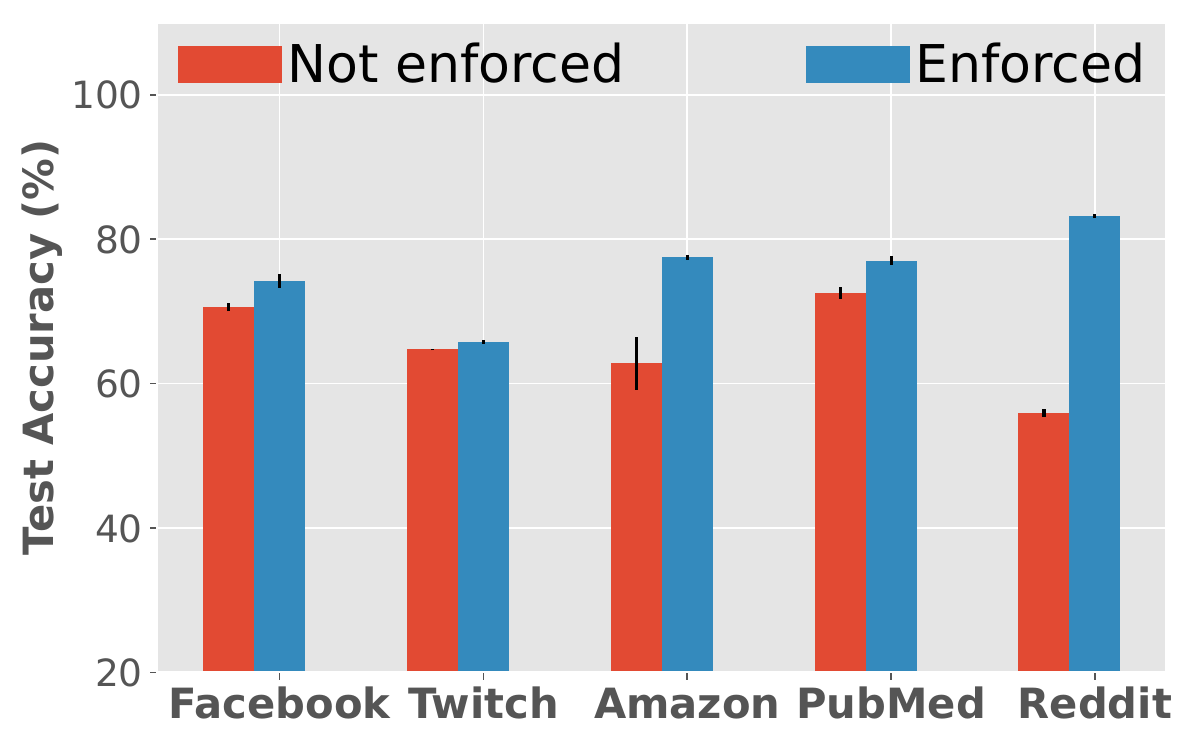}
    \label{fig:no_overlap_2}}
    \subfloat[$\varepsilon=16$]{\includegraphics[width=.49\linewidth]{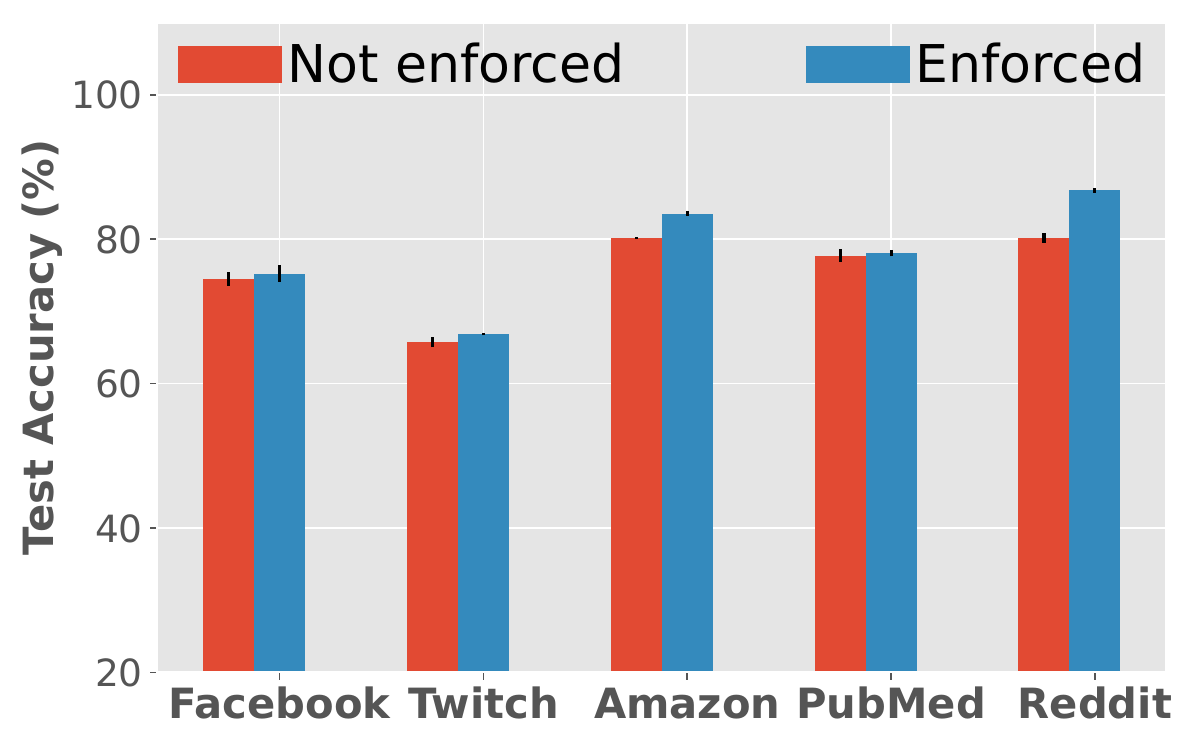}
    \label{fig:no_overlap_16}}\\

    \caption{Performance comparison between when line \ref{alg:HeterPoisson_overlapping_s} to \ref{alg:HeterPoisson_overlapping_e} in Algorithm \ref{alg:HeterPoisson} is ``Enforced'' and ``Not enforced''.}
    \label{fig:no_overlap} 
    
\end{figure}

\vspace{0.1cm}
\noindent\textbf{Necessity of enforcing no \textit{peripheral} node overlaps with \textit{central} nodes}. We study the effect of line \ref{alg:HeterPoisson_overlapping_s} to \ref{alg:HeterPoisson_overlapping_e} in Algorithm \ref{alg:HeterPoisson}. In principle, without such operations, our privacy accounting can still apply. 
It is easy to find the new distribution, as elaborated below: 

we consider two separate cases: 1) the differing node is sampled as a \textit{central} node, then $$\operatorname{Pr}(k+0.5)=q_b\mathbf{Bi}(k; D_{ot}, \frac{q_bM}{D_{ot}}) \text{ for } k\in[D_{ot}],$$
and 
$$\operatorname{Pr}(k)=(1-q_b)\mathbf{Bi}(k; D_{ot}, \frac{q_bM}{D_{ot}}) \text{ for } k\in[D_{ot}].$$ However, in total, this increases the upper bound in our RDP bound, resulting in larger noise s.t.d. to ensure the same $(\varepsilon,\delta)$-DP. Hence, it potentially hurts utility. 

 We present the results for cases when Line \ref{alg:HeterPoisson_overlapping_s} to \ref{alg:HeterPoisson_overlapping_e} is \textit{enforce} compared to \textit{not enforced} in Figure \ref{fig:no_overlap}. The results show that the \textit{enforced} version performs better due to smaller noise s.t.d., especially under the most private case. This suggests that although the enforcing line \ref{alg:HeterPoisson_overlapping_s} to \ref{alg:HeterPoisson_overlapping_e} will alter the sampled sub-graphs, it still benefits utility, suggesting that in real-world applications, line \ref{alg:HeterPoisson_overlapping_s} to \ref{alg:HeterPoisson_overlapping_e} in Algorithm \ref{alg:HeterPoisson} is necessary.

\begin{figure}[!ht] 
    \centering
    \subfloat[$\varepsilon=2$]
    {\includegraphics[width=1\linewidth]{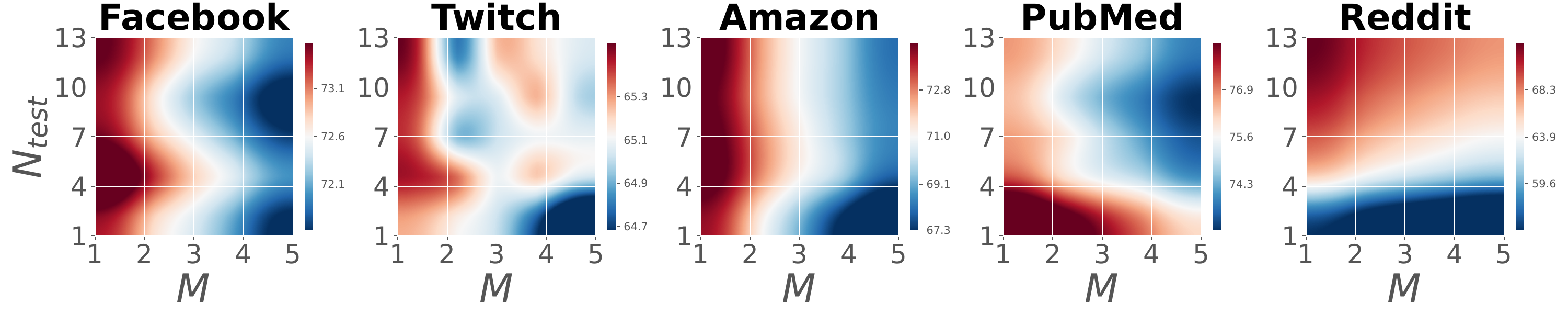}
    \label{fig:trade_offs_ns_gcn_eps1}}\\
    \subfloat[$\varepsilon=\infty$]{\includegraphics[width=1\linewidth]{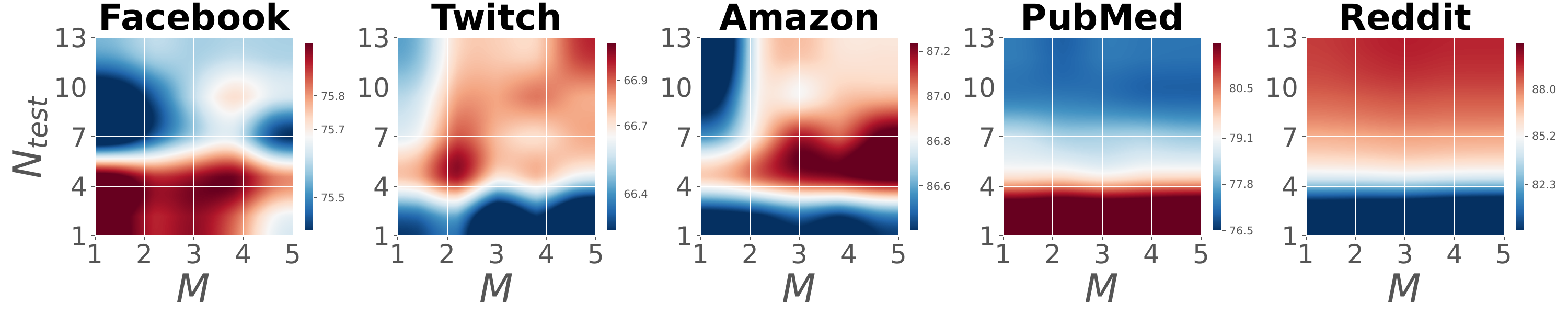}
    \label{fig:trade_offs_ns_gcn_eps10000}}\\
    
    \caption{Trade-offs in neighborhood sampling in transductive graph setting. We interpolate the accuracy value to make it easier to see the trend. Closer to dark red means better performance.}
    \label{fig:trade_offs_ns_gcn} 
    
\end{figure}

\vspace{0.1cm}
\noindent\textbf{Trade-offs in neighborhood sampling}. Increasing $M$ leads to aggregating more information from neighbors, which is good for utility; however, increasing $M$ will also increase the sampling rate, hence less private. In other words, it needs more DP noise to guarantee the same privacy level. Therefore, a trade-off exists in selecting $M$. In addition to $M$, we use $N_{test}$ to denote the number of neighbors to sample during testing, and it is another quantity worthy of investigation.

Results corresponding to GCN model are presented in Figure \ref{fig:trade_offs_ns_gcn}. We give results on privacy levels ($\varepsilon=2,\infty$). As shown in Figure \ref{fig:trade_offs_ns_gcn}, in the most private case, there is an obvious \textit{noise-averse} phenomenon, {\it i.e.}, although larger $M$ means \textit{central} nodes aggregating more information from neighborhoods, to guarantee the same privacy level, the additional noise needed outweighs such merit. 
In contrast, when no noise is added ($\varepsilon=\infty$), we can see that a larger $M$ is favored, which indeed shows that aggregating more neighbors benefits utility. The above phenomenon indicates that the DP noise has a significant impact on the utility. This suggests that, in practice, some small $M$ value is favored during training. From these results, we can also conclude that larger $N_{test}$ tends to benefit utility. However, the experimental results on dataset PubMed show that the best $N_{test}$ lies between 1-4 instead of some larger value. Possibly, this is because the dataset Pubmed itself favors sampling a smaller number of neighbors, as its average edges (only 5) are considerably less than that of the rest of the datasets.

\begin{figure*}[!ht] 
    \centering
    \subfloat[Attack accuracy under different $\varepsilon$, ours]
    {\includegraphics[width=.5\linewidth]{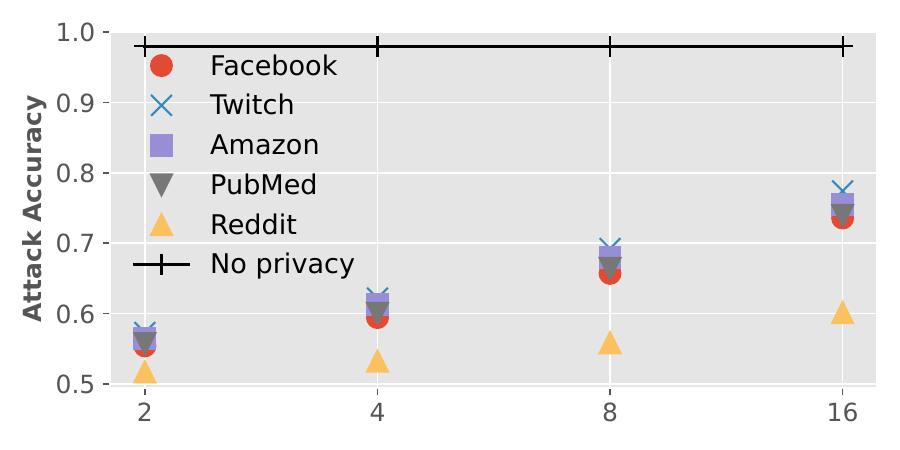}
    \label{fig:audit_acc_our}}
    \subfloat[Empirical $\varepsilon$ by privacy audit, ours]{\includegraphics[width=.5\linewidth]{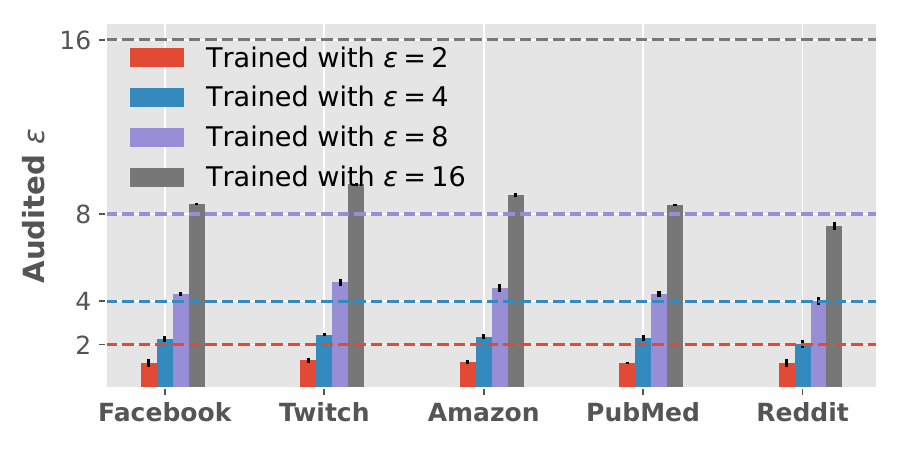}
    \label{fig:audit_audit_our}}\\
    \subfloat[Attack accuracy under different $\varepsilon$, NDP]
    {\includegraphics[width=.5\linewidth]{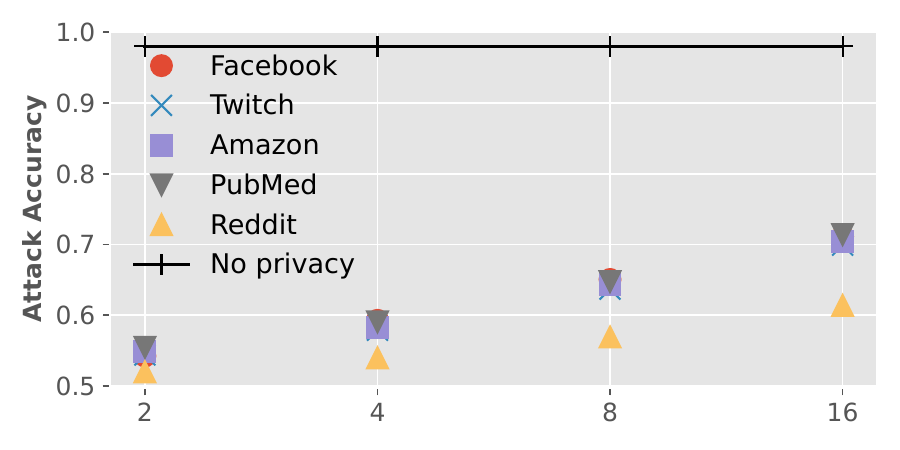}
    \label{fig:audit_acc_ndp}}
    \subfloat[Empirical $\varepsilon$ by privacy audit, NDP]{\includegraphics[width=.5\linewidth]{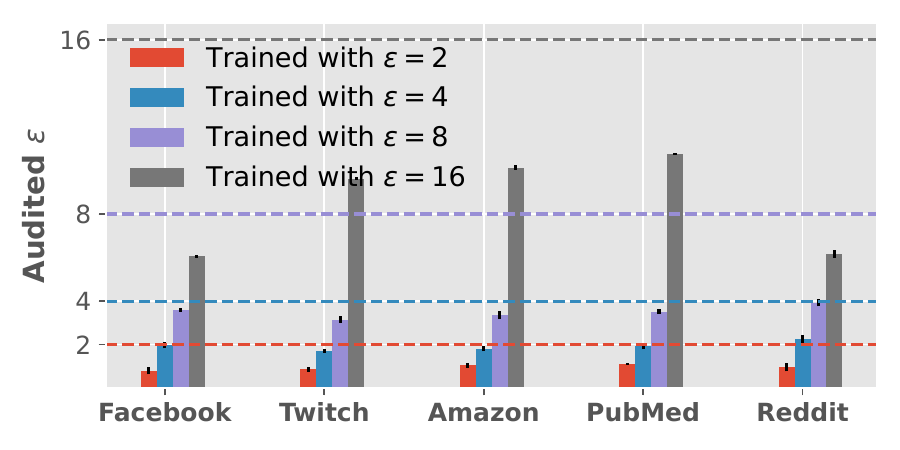}
    \label{fig:audit_audit_ndp}}\\

    \caption{Results of private attack/audit on our protocol. In (a), we present the membership inference attacks on our protocol. In (b), we present the privacy audit results with 95\% confidence. Experiments are performed under a strong adversary model, {\it i.e.}, the adversary has access to all intermediate privatized gradient vectors and can insert gradient canaries. }
    \label{fig:audit} 
    
\end{figure*}

\subsection{Privacy Audit \& Resistance to Privacy Attacks}

Privacy audit focusing on the validation of theoretical privacy guarantees. In this part, we apply the privacy audit method in \cite{DBLP:conf/uss/NasrH0BTJCT23_audits,DBLP:conf/sp/NasrSTPC21} to demonstrate that our protocol indeed provides claimed DP guarantees. To test our method in a real-world scenario with privacy threats, we also apply the strong adversary model assumed by privacy auditing to perform a membership inference attack, and this is to test the limit of our method and showcase our protocol's resistance to (even strong) privacy attacks. The algorithm for auditing adapted from \cite{DBLP:conf/uss/NasrH0BTJCT23_audits}, is provided in Algorithm \ref{alg:privacy_audit} of Appendix \ref{app:privacy_audit}.

We perform such evaluation on both our method (Figure \ref{fig:audit_acc_our} and \ref{fig:audit_audit_our}) and NDP \cite{DBLP:journals/corr/abs-2111-15521-node-level_deva} (Figure \ref{fig:audit_acc_ndp} and \ref{fig:audit_audit_ndp}). 
In Figure \ref{fig:audit_acc_our}, we can see that under the most private scenario ($\varepsilon=2$), the attack accuracy is close to random guessing. As we have lower privacy requirements, naturally, we observe an increase in attack accuracy. Similar attack performance is observed when evaluated on NDP in Figure \ref{fig:audit_acc_ndp}, and results suggest that our method's defense against privacy attack is roughly the same as that of NDP. 

In Figure \ref{fig:audit_audit_our}, we show the audit results on our protocol. We observe that the audited/empirical $\varepsilon$ is close to the theoretical value when $\varepsilon=2$; however, as $\varepsilon$ increases, a wider gap is observed. The fact that no audited/empirical $\varepsilon$ is higher than the theoretical value suggests that our protocol is private as claimed. We may also conclude that when we are at a higher privacy regime, our protocol's DP guarantee becomes tighter. We can also see that the privacy accounting of NDP has no bugs, as shown in Figure \ref{fig:audit_audit_ndp}.

\section{Study on Private Node Embedding}\label{sec:private_node_embedding}

The purpose of presenting this section is not only to show that \cite{sajadmanesh2023gap} fails to ensure claimed privacy protection but also to show that the private node embedding approach fundamentally cannot achieve good utility in high privacy regimes. This implication is important as it is not limited to only graph data. We first present the definition of private node embedding as follows.

\begin{definition}[Private Node Embedding \cite{sajadmanesh2023gap}]\label{def:private_embedding}
    For a graph $\mathcal{G}^*=(\mathcal{V}, \mathcal{E}, \mathbf{X}, \mathbf{Y})$, the private node embeddings (with dimension $k$) for each node are derived by a sequence of private algorithms $\mathcal{M} = (\mathcal{M}_1, \mathcal{M}_2, \cdots, \mathcal{M}_{|\mathcal{V}|})$ such that $\mathcal{M}_i(\mathcal{G})\in \mathbb{R}^k$ outputs the embedding for node $i$ and $\mathcal{M}_i$ is $(\varepsilon_i, \delta_i)$-DP for $i\in\mathcal{V}$, {\it i.e.}, 
    \begin{equation}\label{equ:private_node_embedding}
    \begin{aligned}
        \operatorname{Pr} (\mathcal{M}_i(\mathcal{G}^*)\in S) \leq e^{\varepsilon_i} \operatorname{Pr}(\mathcal{M}_i(\mathcal{G'})\in S) + \delta_i
     \end{aligned}
    \end{equation}
    holds for any $S\subseteq \mathbb{R}^k$ and any adjacent graph pair $(\mathcal{G}^*, \mathcal{G'})$.
\end{definition}

\subsection{Impossibility Result}

\begin{definition}[Class-$\zeta$-aligned Classifier]\label{def:align_classifier}
 Given any graph $\mathcal{G}=(\mathcal{V}, \mathcal{E}, \mathbf{X}, \mathbf{Y})$, and private node embeddings algorithm $\mathcal{M}$. Let node $i$ be uniformly randomly chosen from the node set $\mathcal{V}$, and then we derive its private embedding $u_i=\mathcal{M}_i(\mathcal{G})$, i.e., $u_i$ is a random vector, and randomness comes from both sampling $i$ and the private node embedding algorithm $\mathcal{M}_i$. A classifier $h$ is said to be Class-$\zeta$-Aligned if
\begin{equation}\label{equ:align_classifier}
    \begin{aligned}
        \operatorname{Pr}(\mathbf{Y}[i]=\zeta) =\operatorname{Pr}(h(u_i)=\zeta).
    \end{aligned}
    \end{equation}
\end{definition}
Intuitively, Equation \eqref{equ:align_classifier} says that the portion of nodes whose class is $\zeta$ is equal to 
$$\frac{\# \text{ nodes predicted to be class } \zeta}{\# \text{ all predictions }}.$$ It is easy to see that, for a good classifier predicting the class of node $i$, Equation \eqref{equ:align_classifier} should hold, or at least $h$ should satisfy $\operatorname{Pr}(\mathbf{Y}[i]=\zeta) \approx \operatorname{Pr}(h(u_i)=\zeta)$. We assume such a good classifier to show that \textit{even if we have such a good one, a utility barrier still exists}.

For simplicity and ease of discussion, we take $\operatorname{Pr}(\mathbf{Y}[i]=\zeta) = \operatorname{Pr}(h(u_i)=\zeta)$. Based on Definition \ref{def:align_classifier}, the following theorem shows that there exists a utility barrier for private node embedding.

\begin{theorem}[Impossibility Result]\label{thm:3}\label{thm:private_embedding_impossible_special}
Consider a graph $\mathcal{G}$ whose nodes have $C$ classes. We have a sequence of private node embeddings $\mathcal{M} = (\mathcal{M}_1, \mathcal{M}_2,\cdots, \mathcal{M}_{|\mathcal{V}|})$ where each $\mathcal{M}_k$ is $(\varepsilon, \delta)$-DP for $k\in\{1,2,\cdots, |\mathcal{V}|\}$. Then, for an arbitrary classifier $h$ which is Class-$\zeta$-Aligned for some $\zeta\in \{1,2,\cdots, C\}$, we have the following bound on the precision of $h$ if $h$ operates on $u_i$, which is defined in Definition \ref{def:align_classifier}  
\begin{equation}\label{equ:precision_upper_bound}
\begin{aligned}
    \operatorname{Pr}(\mathbf{Y}[i]=\zeta | h(u_i)=\zeta) \leq \frac{e^{\varepsilon}+\delta(C-1)}{C-1+e^{\varepsilon}}. 
\end{aligned}
\end{equation}
\end{theorem}
Proof is provided in Appendix \ref{appendix:proof_private_node_embedding_special}. From Equation \eqref{equ:precision_upper_bound}, we can easily see that when in the high privacy regime, {\it i.e.}, $\varepsilon \rightarrow 0$, the upper bound is close to $\frac{1}{C}$. This is close to random guessing as $\mathcal{M}_i$ outputs a purely random vector in the extreme case. Note that there is no assumption on the classifier $h$ other than it is Class-$\zeta$-Aligned, so Equation \eqref{equ:precision_upper_bound} holds for any classifier, including potentially well-trained ones that perform optimally on non-private test data. The impossibility result shows that it is impossible to have private $\mathcal{M}_i$ and high utility from the output of $\mathcal{M}_i$ simultaneously. For example, when we have $C=10$ classes in total and each $\mathcal{M}_i$ is $(2,10^{-5})$-DP, the precision upper bound is almost $\frac{e^2}{9+e^2}\approx 0.45$.

\begin{table*}[!ht] 
\begin{subtable}[h]{\textwidth}
    \raggedleft
    \resizebox{\columnwidth}{3.1cm}{
    \begin{tabular}{l*{15}{|c}}
    \toprule
    \multirow{2}{*}{{\Large $\varepsilon$}} & \multicolumn{5}{c|}{GCN} & \multicolumn{5}{c|}{GIN} & \multicolumn{5}{c}{SAGE}\\
    & \textbf{Facebook} & \textbf{Twitch} & \textbf{Amazon} & \textbf{PubMed} & \textbf{Reddit} & \textbf{Facebook} & \textbf{Twitch} & \textbf{Amazon} & \textbf{PubMed} & \textbf{Reddit} & \textbf{Facebook} & \textbf{Twitch} & \textbf{Amazon} & \textbf{PubMed} & \textbf{Reddit} \\
    
\midrule
\multirow{5}{*}{2}&{\large $(71.1)$}&{\large $(88.1)$}&{\large $(45.1)$}&{\large $(78.7)$}&{\large $(15.6)$}&{\large $-$}&{\large $-$}&{\large $-$}&{\large $-$}&{\large $-$}&{\large $-$}&{\large $-$}&{\large $-$}&{\large $-$}&{\large $-$}\\
&{{\large ${49.2}_{\pm 2.0}$}}&{{\large ${42.5}_{\pm 8.2}$}}&{{\large ${14.7}_{\pm 0.4}$}}&{{\large ${15.4}_{\pm 0.4}$}}&{{\large ${32.4}_{\pm 1.6}$}}&{\large $-$}&{\large $-$}&{\large $-$}&{\large $-$}&{\large $-$}&{\large $-$}&{\large $-$}&{\large $-$}&{\large $-$}&{\large $-$}\\
&{\large $34.3_{\pm 0.9}$}&{\large $55.9_{\pm 0.3}$}&{\large $34.9_{\pm 0.9}$}&{\large $41.0_{\pm 1.2}$}&{\large $16.0_{\pm 0.3}$}&{\large $-$}&{\large $-$}&{\large $-$}&{\large $-$}&{\large $-$}&{\large $-$}&{\large $-$}&{\large $-$}&{\large $-$}&{\large $-$}\\
&{\large $21.8_{\pm 0.1}$}&{\large $17.9_{\pm 0.5}$}&{\large $27.4_{\pm 0.4}$}&{\large $42.6_{\pm 0.3}$}&{\large $43.5_{\pm 0.3}$}&{\large $36.9_{\pm 0.9}$}&{\large $7.3_{\pm 0.9}$}&{\large $9.1_{\pm 0.9}$}&{\large $34.7_{\pm 0.5}$}&{\large $35.8_{\pm 0.3}$}&{\large $22.5_{\pm 0.6}$}&{\large $4.8_{\pm 0.4}$}&{\large $25.2_{\pm 0.9}$}&{\large $44.3_{\pm 0.9}$}&{\large $39.1_{\pm 0.4}$}\\
&\cellcolor{gray!30}\underline{{\large $\textbf{74.1}_{\pm 1.1}$}}&\cellcolor{gray!30}\underline{{\large $\textbf{64.8}_{\pm 0.2}$}}&\cellcolor{gray!30}\underline{{\large $\textbf{75.3}_{\pm 0.9}$}}&\cellcolor{gray!30}\underline{{\large $\textbf{78.0}_{\pm 0.3}$}}&\cellcolor{gray!30}\underline{{\large $\textbf{82.9}_{\pm 0.1}$}}&\cellcolor{gray!30}\underline{{\large $\textbf{74.7}_{\pm 0.9}$}}&\cellcolor{gray!30}\underline{{\large $\textbf{64.5}_{\pm 0.4}$}}&\cellcolor{gray!30}\underline{{\large $\textbf{74.4}_{\pm 0.8}$}}&\cellcolor{gray!30}\underline{{\large $\textbf{82.3}_{\pm 0.6}$}}&\cellcolor{gray!30}\underline{{\large $\textbf{80.6}_{\pm 0.4}$}}&\cellcolor{gray!30}\underline{{\large $\textbf{74.2}_{\pm 0.4}$}}&\cellcolor{gray!30}\underline{{\large $\textbf{64.9}_{\pm 0.5}$}}&\cellcolor{gray!30}\underline{{\large $\textbf{73.9}_{\pm 0.8}$}}&\cellcolor{gray!30}\underline{{\large $\textbf{80.8}_{\pm 0.5}$}}&\cellcolor{gray!30}\underline{{\large $\textbf{80.2}_{\pm 0.4}$}}\\
\midrule
\multirow{4}{*}{4}&{{\large ${49.4}_{\pm 2.2}$}}&{{\large ${40.9}_{\pm 8.8}$}}&{{\large ${14.4}_{\pm 0.9}$}}&{{\large ${15.4}_{\pm 0.4}$}}&{{\large ${33.3}_{\pm 1.5}$}}&{\large $-$}&{\large $-$}&{\large $-$}&{\large $-$}&{\large $-$}&{\large $-$}&{\large $-$}&{\large $-$}&{\large $-$}&{\large $-$}\\
&{\large $35.2_{\pm 0.4}$}&{\large $59.0_{\pm 0.4}$}&{\large $30.9_{\pm 0.6}$}&{\large $34.8_{\pm 0.4}$}&{\large $16.4_{\pm 0.9}$}&{\large $-$}&{\large $-$}&{\large $-$}&{\large $-$}&{\large $-$}&{\large $-$}&{\large $-$}&{\large $-$}&{\large $-$}&{\large $-$}\\
&{\large $51.4_{\pm 0.2}$}&{\large $35.9_{\pm 0.4}$}&{\large $31.1_{\pm 0.9}$}&{\large $40.7_{\pm 0.5}$}&{\large $54.3_{\pm 0.9}$}&{\large $49.0_{\pm 0.4}$}&{\large $35.1_{\pm 0.9}$}&{\large $16.8_{\pm 0.9}$}&{\large $41.0_{\pm 0.4}$}&{\large $47.7_{\pm 0.9}$}&{\large $39.9_{\pm 0.3}$}&{\large $19.0_{\pm 0.5}$}&{\large $24.5_{\pm 0.2}$}&{\large $41.2_{\pm 0.9}$}&{\large $50.0_{\pm 0.2}$}\\
&\cellcolor{gray!30}\underline{{\large $\textbf{74.6}_{\pm 1.1}$}}&\cellcolor{gray!30}\underline{{\large $\textbf{65.6}_{\pm 0.5}$}}&\cellcolor{gray!30}\underline{{\large $\textbf{78.6}_{\pm 1.0}$}}&\cellcolor{gray!30}\underline{{\large $\textbf{78.7}_{\pm 0.1}$}}&\cellcolor{gray!30}\underline{{\large $\textbf{84.7}_{\pm 0.2}$}}&\cellcolor{gray!30}\underline{{\large $\textbf{75.8}_{\pm 0.6}$}}&\cellcolor{gray!30}\underline{{\large $\textbf{65.2}_{\pm 1.1}$}}&\cellcolor{gray!30}\underline{{\large $\textbf{78.3}_{\pm 1.3}$}}&\cellcolor{gray!30}\underline{{\large $\textbf{84.0}_{\pm 0.4}$}}&\cellcolor{gray!30}\underline{{\large $\textbf{82.9}_{\pm 0.1}$}}&\cellcolor{gray!30}\underline{{\large $\textbf{74.9}_{\pm 0.6}$}}&\cellcolor{gray!30}\underline{{\large $\textbf{65.7}_{\pm 0.4}$}}&\cellcolor{gray!30}\underline{{\large $\textbf{79.1}_{\pm 1.7}$}}&\cellcolor{gray!30}\underline{{\large $\textbf{81.7}_{\pm 0.3}$}}&\cellcolor{gray!30}\underline{{\large $\textbf{82.7}_{\pm 0.3}$}}\\
\midrule
\multirow{4}{*}{8}&{{\large ${49.5}_{\pm 1.8}$}}&{{\large ${40.9}_{\pm 8.8}$}}&{{\large ${14.7}_{\pm 1.5}$}}&{{\large ${15.4}_{\pm 0.4}$}}&{{\large ${35.0}_{\pm 1.9}$}}&{\large $-$}&{\large $-$}&{\large $-$}&{\large $-$}&{\large $-$}&{\large $-$}&{\large $-$}&{\large $-$}&{\large $-$}&{\large $-$}\\
&{\large $34.1_{\pm 1.2}$}&{\large $56.7_{\pm 0.9}$}&{\large $36.1_{\pm 0.9}$}&{\large $33.3_{\pm 0.3}$}&{\large $19.7_{\pm 0.4}$}&{\large $-$}&{\large $-$}&{\large $-$}&{\large $-$}&{\large $-$}&{\large $-$}&{\large $-$}&{\large $-$}&{\large $-$}&{\large $-$}\\
&{\large $60.8_{\pm 0.4}$}&{\large $65.7_{\pm 0.2}$}&{\large $30.9_{\pm 0.6}$}&{\large $41.7_{\pm 0.3}$}&{\large $67.1_{\pm 0.6}$}&{\large $69.2_{\pm 0.4}$}&{\large $43.5_{\pm 0.9}$}&{\large $29.9_{\pm 0.9}$}&{\large $49.9_{\pm 0.4}$}&{\large $58.4_{\pm 0.4}$}&{\large $52.8_{\pm 0.3}$}&{\large $58.3_{\pm 0.3}$}&{\large $34.9_{\pm 0.9}$}&{\large $41.7_{\pm 0.4}$}&{\large $64.2_{\pm 0.5}$}\\
&\cellcolor{gray!30}\underline{{\large $\textbf{74.7}_{\pm 1.2}$}}&\cellcolor{gray!30}\underline{{\large $\textbf{65.6}_{\pm 0.3}$}}&\cellcolor{gray!30}\underline{{\large $\textbf{81.1}_{\pm 0.9}$}}&\cellcolor{gray!30}\underline{{\large $\textbf{79.4}_{\pm 0.3}$}}&\cellcolor{gray!30}\underline{{\large $\textbf{85.9}_{\pm 0.3}$}}&\cellcolor{gray!30}\underline{{\large $\textbf{76.6}_{\pm 0.5}$}}&\cellcolor{gray!30}\underline{{\large $\textbf{65.3}_{\pm 0.5}$}}&\cellcolor{gray!30}\underline{{\large $\textbf{81.6}_{\pm 1.3}$}}&\cellcolor{gray!30}\underline{{\large $\textbf{84.8}_{\pm 0.3}$}}&\cellcolor{gray!30}\underline{{\large $\textbf{84.4}_{\pm 0.1}$}}&\cellcolor{gray!30}\underline{{\large $\textbf{75.4}_{\pm 0.7}$}}&\cellcolor{gray!30}\underline{{\large $\textbf{65.9}_{\pm 0.1}$}}&\cellcolor{gray!30}\underline{{\large $\textbf{82.4}_{\pm 0.7}$}}&\cellcolor{gray!30}\underline{{\large $\textbf{82.4}_{\pm 0.0}$}}&\cellcolor{gray!30}\underline{{\large $\textbf{84.1}_{\pm 0.3}$}}\\
\midrule
\multirow{4}{*}{16}&{{\large ${49.8}_{\pm 1.3}$}}&{{\large ${40.8}_{\pm 8.8}$}}&{{\large ${15.1}_{\pm 2.1}$}}&{{\large ${15.4}_{\pm 0.4}$}}&{{\large ${35.8}_{\pm 2.0}$}}&{\large $-$}&{\large $-$}&{\large $-$}&{\large $-$}&{\large $-$}&{\large $-$}&{\large $-$}&{\large $-$}&{\large $-$}&{\large $-$}\\
&{\large $32.6_{\pm 0.1}$}&{\large $56.9_{\pm 0.9}$}&{\large $34.6_{\pm 1.2}$}&{\large $31.4_{\pm 0.4}$}&{\large $24.5_{\pm 1.2}$}&{\large $-$}&{\large $-$}&{\large $-$}&{\large $-$}&{\large $-$}&{\large $-$}&{\large $-$}&{\large $-$}&{\large $-$}&{\large $-$}\\
&{\large $73.8_{\pm 0.1}$}&{\large $68.1_{\pm 0.2}$}&{\large $45.5_{\pm 0.9}$}&{\large $43.2_{\pm 0.9}$}&{\large $71.7_{\pm 0.3}$}&{\large $77.4_{\pm 0.9}$}&{\large $52.4_{\pm 0.9}$}&{\large $40.0_{\pm 0.9}$}&{\large $54.8_{\pm 0.9}$}&{\large $61.8_{\pm 0.9}$}&{\large $69.0_{\pm 0.1}$}&{\large $65.6_{\pm 0.4}$}&{\large $37.5_{\pm 0.3}$}&{\large $43.2_{\pm 0.9}$}&{\large $73.8_{\pm 0.9}$}\\
&\cellcolor{gray!30}\underline{{\large $\textbf{75.1}_{\pm 1.2}$}}&\cellcolor{gray!30}\underline{{\large $\textbf{66.0}_{\pm 0.2}$}}&\cellcolor{gray!30}\underline{{\large $\textbf{82.1}_{\pm 0.9}$}}&\cellcolor{gray!30}\underline{{\large $\textbf{79.5}_{\pm 0.4}$}}&\cellcolor{gray!30}\underline{{\large $\textbf{86.8}_{\pm 0.3}$}}&\cellcolor{gray!30}\underline{{\large $\textbf{76.6}_{\pm 0.8}$}}&\cellcolor{gray!30}\underline{{\large $\textbf{65.8}_{\pm 0.2}$}}&\cellcolor{gray!30}\underline{{\large $\textbf{82.7}_{\pm 2.0}$}}&\cellcolor{gray!30}\underline{{\large $\textbf{85.5}_{\pm 0.3}$}}&\cellcolor{gray!30}\underline{{\large $\textbf{85.5}_{\pm 0.1}$}}&\cellcolor{gray!30}\underline{{\large $\textbf{75.5}_{\pm 0.7}$}}&\cellcolor{gray!30}\underline{{\large $\textbf{66.1}_{\pm 0.4}$}}&\cellcolor{gray!30}\underline{{\large $\textbf{83.2}_{\pm 1.4}$}}&\cellcolor{gray!30}\underline{{\large $\textbf{82.9}_{\pm 0.1}$}}&\cellcolor{gray!30}\underline{{\large $\textbf{85.3}_{\pm 0.2}$}}\\

\bottomrule
\end{tabular}
    }

\end{subtable}

\caption{Classification precision (averaged across classes). The organization of the presentation is identical to Table \ref{tab:compare_with_GAP_and_deva}. We add the precision upper bound, {\it i.e.}, Equation \eqref{equ:precision_upper_bound}, in the first row (in parenthesis) for $\varepsilon=2$. For larger $\varepsilon$, we omit the upper bound as it becomes trivial.}
\label{tab:compare_with_GAP_and_deva_precision}

\end{table*}

\vspace{0.2cm}
\noindent\textbf{Implications}. The impossibility result reveals that releasing private node embeddings for each node in a graph has poor trade-offs between privacy and utility. This phenomenon coincides with one of the limitations of DP: \textit{differential privacy is not a meaningful privacy notion if the analysis action is taken on a specific individual} \cite{DBLP:books/sp/17/Vadhan17_comp_DP}. Our impossibility results are linked to such a claim. Unfortunately, private node embedding falls into this case. It can be more intuitive by considering the following: Naturally, we want the embeddings of two totally different nodes/instances to be \textit{distinguishable} to each other; however, in privacy node/instance embedding, under DP's definition, it enforces the distributions of such two embeddings to be \textit{close/indistinguishable} to each other, conflicting with our expectation, hence not possible to have both strong privacy and good utility. 

We provide a case study in Section \ref{sec:case_study} that evaluates a method using private node embedding \cite{sajadmanesh2023gap}. The analysis of such a method on privacy has flaws, resulting in a significantly underestimated amount of DP noise, which leads to incorrect privacy guarantees.

\vspace{0.1cm}
\subsection{Case study}\label{sec:case_study}

\noindent\textbf{A fundamental flaw: \textit{Unbounded} or \textit{bounded} DP}? 
Note that the definition of private node embedding is only reasonable under 
the \textit{bounded} DP setting. Suppose there is a pair of two adjacent graphs $\mathcal{G}^*$ and $\mathcal{G'}=\mathcal{G}^*\cup \{\text{node } z\}$, if the \textit{unbounded} DP is adopted as its privacy definition, there is a problem when writing down 
\begin{equation}\nonumber
\operatorname{Pr} (\mathcal{M}_z(\mathcal{G'})\in S) \leq e^{\varepsilon} \operatorname{Pr}(\mathcal{M}_z(\mathcal{G}^*)\in S) + \delta,
\end{equation}
{\it i.e.}, as there is no node $z$ in $\mathcal{G}^*$. Therefore, this type of information publishing in privacy node/instance embedding is incompatible with \textit{Unbounded} DP formulation.

Besides the above technical explanation, considering the following from an adversary point of view can be much more intuitive. Since the embedding is already claimed private, the privacy adversary can access those private outputs. Suppose we have $|\mathcal{V}|$ nodes in $\mathcal{G}^*$, the adversary can just count the number ($|\mathcal{V}|$ V.S. $|\mathcal{V}|+1$) of embeddings to perfectly distinguish the graph from which those embeddings are derived, thus perfectly infers whether $z$ has participated or not. Note that the adversary will always succeed no matter how much noise is added to the embeddings. Unfortunately, we notice the private node embedding use case with \textit{unbounded} DP formulation \cite{sajadmanesh2023gap}.

\vspace{0.1cm}
\noindent\textbf{Incorrect sensitivity analysis}. We know that private node embedding is only compatible with \textit{bounded} DP formulation, and the following discussion is carried out based on this. In \cite{sajadmanesh2023gap}, a private node embedding for node $i$ needs to be released. Simply speaking, DP noise is added to the result 
\begin{equation}\nonumber
    \hat{x}_i=\frac{x_i}{\|x_i\|}+\sum_{j\in\mathcal{NB}(i)}\frac{x_j}{\|x_j\|}
\end{equation}
for each node $i$ ($x_j$ is the feature vector, and the normalization operation is to enforce bounded sensitivity).

\cite{sajadmanesh2023gap} analyzes that the $\ell_2$ sensitivity of $x'_i$ is 1 by Lemma 2 of \cite{sajadmanesh2023gap}, and it corresponds to the case when one of $i$'s neighbors is the differing node. However, this analysis is problematic because the differing node can be $i$ itself, and the differing node can differ in all of its neighbors, hence the sensitivity is substantially underestimated, specifically when the nodes have $\mathrm{D}$ degrees in maximum, the DP noise is underestimated by a $2(\mathrm{D}+1)$ factor (by definition, the $\ell_2$ sensitivity is the $\ell_2$ norm of the difference between the sum of arbitrary $(\mathrm{D}+1)$ normalized vectors and the sum of another $(\mathrm{D}+1)$ normalized vectors, such norm is $2(\mathrm{D}+1)$ in maximum by triangle inequality). Therefore, to correct the amount of DP noise, the noise s.t.d. in \cite{sajadmanesh2023gap} should have been multiplied by $2(\mathrm{D}+1)$, the flaw in Lemma 2 in \cite{sajadmanesh2023gap} propagates to Lemma 3 and the main theorem in \cite{sajadmanesh2023gap}. By this, one can see that there is little utility remaining as the noise intensity always scales with the size of neighbors of one node. Our impossibility results shown previously capture this phenomenon. 

\subsection{Experiments on Precision}

With the same experiment setup as in Section \ref{sec:eval}, we add the precision results to confirm our precision upper bound in Equation \eqref{equ:precision_upper_bound}. Results under the transductive setting are presented in Table \ref{tab:compare_with_GAP_and_deva_precision}. We can see that the performance of GAP \cite{sajadmanesh2023gap} is below the upper bound after the random noise is corrected, confirming the utility barrier. In contrast, for other methods, it can be seen that they perform better than the upper bound naturally because other methods do not use private instance embedding, hence not limited to the utility barrier. Note that as we have less privacy ($\varepsilon$ becomes greater), the precision upper bound becomes trivial (approaching 1), which is the reason why we do not list the bound for settings with lower privacy in Table \ref{tab:compare_with_GAP_and_deva_precision}.

\section{Related Work}\label{sec:related_work}

\noindent\textbf{Differential privacy in machine learning}. Depending on where the randomness is injected, there are roughly three categories: 1) \textit{output pertubation}, this method perturb the trained model \cite{DBLP:conf/uss/Jayaraman019, DBLP:conf/nips/ChaudhuriM08}
; 2) \textit{object modification}, this method adds randomness to the loss function \cite{DBLP:journals/pvldb/ZhangZXYW12,DBLP:journals/jmlr/KiferST12}
; 3) \textit{gradient perturbation}, this method adds noise to the gradient during training \cite{abadi2016deep, DBLP:conf/focs/BassilyST14}
. Notably, the third one does not assume any convexity of the loss function, which makes it more suitable for general machine learning/deep learning applications. Among successfully applied protocols, DP-SGD \cite{abadi2016deep,DBLP:conf/globalsip/SongCS13} is a prime example. Our work falls under the third category.

\vspace{0.05cm}
\noindent\textbf{Differential privacy in graph statistic analysis}. This type of work, differing from learning tasks, focuses on privacy issues in analyzing graph data statistics. For instance, a line of work tackling the problems in releasing sub-graph counting \cite{DBLP:conf/uss/ImolaMC21,DBLP:journals/corr/abs-1809-02575}
, some other works focus on degree distributions \cite{DBLP:conf/sigmod/DayLL16, DBLP:conf/icdm/HayLMJ09}. 
Interestingly, \textit{node-level} privacy is also considered significantly harder to obtain than  \textit{edge-level} without incurring notable utility loss even in the central DP model \cite{DBLP:conf/ccs/QinYYKX017}.
We also focus on \textit{node-level} privacy, although the tasks are different.

\vspace{0.05cm}
\noindent\textbf{Differentially privacy in GNNs}. Preserving edges/nodes' privacy is also a natural request in GNNs. \textit{Node-level} privacy is much stronger than \textit{edge-level} privacy, as the former implies the latter. There is notable work tackling \textit{edge-level} privacy protection. Wu et al. propose an edge-level privacy protection method using Laplace Mechanism \cite{DBLP:conf/sp/0011L0022/linkteller}. Kolluri et al. provide another \textit{edge-level} privacy solution \cite{DBLP:conf/ccs/KolluriBHS22}, improving above \cite{DBLP:conf/sp/0011L0022/linkteller}. In contrast, there is no paralleled previous work tackling \textit{node-level} privacy protection with strong results. Like graph statistics analysis tasks, compared with \textit{edge-level}, ensuring \textit{node-level} privacy is also more challenging.

\vspace{0.05cm}
\noindent\textbf{Privacy attacks to learning GNNs on graphs}. This type of work is on the \textit{adversary} side while our work is on the \textit{defender} side. Depending on the applications, multiple privacy attacks exist under various threat models. Zhang et al. \cite{DBLP:conf/uss/000100S022_zhang} provide an empirical study on the privacy issues when releasing graph embeddings, demonstrating risks at releasing. Wu et al. \cite{DBLP:conf/sp/0011L0022/linkteller} provide evidence that the existence of an edge can be inferred by querying on GNN models, showing the privacy risks at test/inference. Recent work by He et al. also studies the membership inference attack on nodes \cite{DBLP:journals/corr/abs-2102-05429_node_mem}, highlighting equal urgency of preserving \textit{node-level} privacy in learning GNN on graphs.


\section{Conclusion}

In this study, we present a novel and non-trivial method to protect \textit{node-level} privacy in learning GNNs on graphs. To provide a better understanding of our design, we first introduce a motivating experiment that leads to our designed sampling routine, namely \textit{HeterPoisson}. 
Additionally, we give the privacy analysis for \textit{HeterPoisson}. 
Our experimental results also illustrate the non-trivial utility afforded by this choice. Furthermore, we theoretically establish that the family of spherical noise can be adopted to obfuscate the output of a function with bounded $\ell_2$-sensitivity, which can be of independent interest. To further evaluate the effectiveness of our privacy solution, we subject it to membership inference attacks and privacy audit techniques. Through experimental analysis, we demonstrate that our protocol is resistant to privacy attacks and it has no bugs.

In addition to our privacy solution, we investigate a method that adopts private node/instance embedding to address the issue of \textit{node-level} privacy. Specifically, we conduct a case study to reveal the fundamental privacy flaws in a recent study that adopts this approach. More importantly, we highlight the inherent utility barriers associated with the private instance embedding approach, providing insights into pitfalls that should be avoided in privacy applications.

\section*{Acknowledgement}

\noindent We thank all anonymous reviewers' constructive feedback. Di Wang and Zihang Xiang were supported by BAS/1/1689-01-01, URF/1/4663-01-01, FCC/1/1976-49-01, RGC/3/4816-01-01, and REI/1/4811-10-01 of King Abdullah University of Science and Technology (KAUST) and KAUST-SDAIA Center of Excellence in Data Science and Artificial Intelligence. Tianhao Wang is supported by CNS-2220433.

\changestart
This version of this manuscript is different from the last version in IEEE S\&P. We have mainly revised section 4.3 (and other related content), as we found out there is a bug in the privacy analysis. Nevertheless, the conclusion derived from our experiment and other claims remains the same. We sincerely thank Professor Xiaokui Xiao and Jianxin Wei for their engagement in this revision.
\changeend

\bibliographystyle{plain}
\bibliography{ref}

\appendices

\section{Content for reference}

\subsection{Used GNN model}\label{appendix:gnns}

The following three GNN models are used in our evaluations.
\begin{itemize}
    \item GCN: the aggregation is a weighted summation over representations of the last layer,
    \begin{equation}\nonumber
    \begin{aligned}
        \mathbf{h}_u^{(k+1)} = \phi^{(k)}(\sum_{j \in \mathcal{NB}(u)\cup\{u\}} \frac{1}{\sqrt{d_j d_u}} \mathbf{h}_j^{(k)})
    \end{aligned}
    \end{equation}
    where $d_u$ is the degree of node $u$.
    
    \item GIN: the update can be described as,
    \begin{equation}\nonumber
    \begin{aligned}
        \mathbf{h}_u^{(k+1)} = \phi^{(k)}(\sum_{j \in \mathcal{NB}(u)}\mathbf{h}_j^{(k)}+(1+\lambda)\mathbf{h}_u^{(k)})
    \end{aligned}
    \end{equation}
    where $\lambda$ is a learnable parameter.
    
    \item SAGE: take the $\operatorname{mean}$ aggregator as an example, the update can be described as,
    \begin{equation}\nonumber
    \begin{aligned}
        \mathbf{h}_u^{(k+1)} = \phi^{(k)}(\operatorname{mean}\{\mathbf{h}_j^{(k)}|j \in \mathcal{NB}(u)\cup\{u\}\})
    \end{aligned}
    \end{equation}
\end{itemize}

\noindent\textbf{GNN training routine}. For example, for the node-classification tasks, the training of a GNN model $w$ can be described as the following. 
\begin{enumerate}
    \item  \textit{Forward propagation}: {\it i.e.}, update $\mathbf{h}_i^k, i \in \mathcal{V}$ until we get the results $\mathbf{h}_i^K, i \in \mathcal{V}$ after $K$-th iteration; feed $\mathbf{h}_i^K, i \in \mathcal{V}$ to a classification module (often a fully connected layer followed by a \textit{softmax} layer) to get the predictions $\hat{y}_i, i\in\mathcal{V}$; compute the loss by some loss metric on $\hat{y}_i$ and the ground-truth label $\mathbf{\mathbf{Y}[i]}$ for $i\in\mathcal{V}$. The total loss for a graph $\mathcal{G}$ is often the averaged loss of all nodes, and we denote it as $f(\mathcal{G};w)$ and $f$ is the loss function.
    \item \textit{Backward propagation}: {\it i.e.}, compute the gradient with respect to the learnable parameter $g=\nabla f(\mathcal{G};w)$.
    \item \textit{Model update}: update $w$ using $g$ by gradient decent.
\end{enumerate}





\subsection{Privacy Audit}\label{app:privacy_audit}

The algorithm for privacy audit is presented in Algorithm \ref{alg:privacy_audit}, which is adapted from \cite{DBLP:conf/uss/NasrH0BTJCT23_audits}. Auditing is done on observations $\mathbf{O^*},\mathbf{O'}$ following section 5.2 of \cite{DBLP:conf/uss/NasrH0BTJCT23_audits}; simply speaking, the goal is trying to distinguish whether the canary gradient is included or not. For the threshold setup, we use various threshold values and choose the one with the strongest audit/attack result. The membership inference attack is also carried out based on $\mathbf{O^*},\mathbf{O'}$. 

\begin{algorithm}[!ht]
\caption{White-box Privacy Audit with Gradient Canaries
}\label{alg:privacy_audit}
\begin{algorithmic}[1]
\small
\renewcommand{\algorithmicrequire}{\textbf{Input:}}
\renewcommand{\algorithmicensure}{\textbf{Output:}}

\Require {
Identical to input of Algorithm \ref{alg:nodedp_training}
}
\State Initialize Observations $\mathbf{O^*}\gets\emptyset$, $\mathbf{O'}\gets\emptyset$
\For{$t=1,2,\cdots,T$}

    \State Run line \ref{alg:nodedp_training_subsample} to \ref{alg:nodedp_training_summation_e} in Algorithm \ref{alg:nodedp_training}
    \State  $\triangleright$ Insert the \textbf{Dirac canary gradient} as defined in \cite{DBLP:conf/uss/NasrH0BTJCT23_audits}
    \State  $\triangleright$ Set canary gradient $g^c= [k,0,0,\cdots,0], k\sim \rho$ where distribution $\rho$ is defined in Theorem \ref{thm:privacy_guarantee}
    \State W.p. $q_b$, choose some gradient $\hat{g}_i$ and let $\hat{g}_i\gets \hat{g}_i+g^c$
    \State $\Bar{g}^t \gets \sum \hat{g}_i$ 
    \State $g^t \gets \Bar{g}^t +\mathcal{N}(0,\sigma\mathbb{I}^d)$ 
    \State $\mathbf{O'}[t]\gets \langle g^t,g^c \rangle$, $\mathbf{O^*}[t]\gets \langle g^t-g^c,g^c \rangle$
    \State $w ^t \gets w ^{t-1} - \eta g^t$ 
\EndFor
\Ensure Observations $\mathbf{O^*},\mathbf{O'}$
\end{algorithmic}
\end{algorithm}

\section{Proofs}

\begin{table*}[!t] 
\begin{subtable}[h]{\textwidth}
    \raggedleft
    \resizebox{\columnwidth}{3.1cm}{
    \begin{tabular}{l*{15}{|c}}
    \toprule
    \multirow{2}{*}{{\Large $\varepsilon$}} & \multicolumn{5}{c|}{GCN} & \multicolumn{5}{c|}{GIN} & \multicolumn{5}{c}{SAGE}\\
    & \textbf{Facebook} & \textbf{Twitch} & \textbf{Amazon} & \textbf{PubMed} & \textbf{Reddit} & \textbf{Facebook} & \textbf{Twitch} & \textbf{Amazon} & \textbf{PubMed} & \textbf{Reddit} & \textbf{Facebook} & \textbf{Twitch} & \textbf{Amazon} & \textbf{PubMed} & \textbf{Reddit} \\
    
\midrule
\multirow{4}{*}{2}&{{\large ${48.6}_{\pm 3.7}$}}&{{\large ${59.0}_{\pm 2.8}$}}&{{\large ${37.4}_{\pm 0.9}$}}&{{\large ${39.3}_{\pm 0.5}$}}&{{\large ${41.9}_{\pm 1.6}$}}&{\large $-$}&{\large $-$}&{\large $-$}&{\large $-$}&{\large $-$}&{\large $-$}&{\large $-$}&{\large $-$}&{\large $-$}&{\large $-$}\\
&{\large $34.3_{\pm 0.9}$}&{\large $55.9_{\pm 0.3}$}&{\large $34.9_{\pm 0.9}$}&{\large $41.0_{\pm 1.2}$}&{\large $25.4_{\pm 0.3}$}&{\large $-$}&{\large $-$}&{\large $-$}&{\large $-$}&{\large $-$}&{\large $-$}&{\large $-$}&{\large $-$}&{\large $-$}&{\large $-$}\\
&{\large $11.3_{\pm 0.1}$}&{\large $11.5_{\pm 0.5}$}&{\large $27.9_{\pm 0.4}$}&{\large $43.8_{\pm 0.3}$}&{\large $38.2_{\pm 0.3}$}&{\large $28.0_{\pm 0.9}$}&{\large $27.2_{\pm 0.9}$}&{\large $7.1_{\pm 0.9}$}&{\large $38.8_{\pm 0.5}$}&{\large $32.5_{\pm 0.3}$}&{\large $34.6_{\pm 0.6}$}&{\large $6.3_{\pm 0.4}$}&{\large $13.5_{\pm 0.9}$}&{\large $38.4_{\pm 0.9}$}&{\large $33.8_{\pm 0.4}$}\\
&\cellcolor{gray!30}\underline{{\large $\textbf{74.2}_{\pm 1.0}$}}&\cellcolor{gray!30}\underline{{\large $\textbf{65.9}_{\pm 0.3}$}}&\cellcolor{gray!30}\underline{{\large $\textbf{77.0}_{\pm 0.5}$}}&\cellcolor{gray!30}\underline{{\large $\textbf{78.1}_{\pm 0.3}$}}&\cellcolor{gray!30}\underline{{\large $\textbf{83.1}_{\pm 0.2}$}}&\cellcolor{gray!30}\underline{{\large $\textbf{74.6}_{\pm 0.9}$}}&\cellcolor{gray!30}\underline{{\large $\textbf{65.4}_{\pm 0.3}$}}&\cellcolor{gray!30}\underline{{\large $\textbf{76.4}_{\pm 0.2}$}}&\cellcolor{gray!30}\underline{{\large $\textbf{82.3}_{\pm 0.4}$}}&\cellcolor{gray!30}\underline{{\large $\textbf{80.8}_{\pm 0.6}$}}&\cellcolor{gray!30}\underline{{\large $\textbf{74.4}_{\pm 0.6}$}}&\cellcolor{gray!30}\underline{{\large $\textbf{66.0}_{\pm 0.9}$}}&\cellcolor{gray!30}\underline{{\large $\textbf{76.5}_{\pm 0.1}$}}&\cellcolor{gray!30}\underline{{\large $\textbf{80.9}_{\pm 0.5}$}}&\cellcolor{gray!30}\underline{{\large $\textbf{80.9}_{\pm 0.5}$}}\\
\midrule
\multirow{4}{*}{4}&{{\large ${51.0}_{\pm 3.6}$}}&{{\large ${60.2}_{\pm 0.8}$}}&{{\large ${37.4}_{\pm 0.9}$}}&{{\large ${39.3}_{\pm 0.5}$}}&{{\large ${42.7}_{\pm 1.5}$}}&{\large $-$}&{\large $-$}&{\large $-$}&{\large $-$}&{\large $-$}&{\large $-$}&{\large $-$}&{\large $-$}&{\large $-$}&{\large $-$}\\
&{\large $35.2_{\pm 0.4}$}&{\large $59.0_{\pm 0.4}$}&{\large $30.9_{\pm 0.6}$}&{\large $34.8_{\pm 0.4}$}&{\large $25.5_{\pm 0.9}$}&{\large $-$}&{\large $-$}&{\large $-$}&{\large $-$}&{\large $-$}&{\large $-$}&{\large $-$}&{\large $-$}&{\large $-$}&{\large $-$}\\
&{\large $45.3_{\pm 0.2}$}&{\large $22.5_{\pm 0.4}$}&{\large $29.2_{\pm 0.9}$}&{\large $40.2_{\pm 0.5}$}&{\large $51.4_{\pm 0.9}$}&{\large $56.3_{\pm 0.4}$}&{\large $40.6_{\pm 0.9}$}&{\large $17.1_{\pm 0.9}$}&{\large $35.5_{\pm 0.4}$}&{\large $43.9_{\pm 0.9}$}&{\large $40.0_{\pm 0.3}$}&{\large $23.8_{\pm 0.5}$}&{\large $30.7_{\pm 0.2}$}&{\large $43.5_{\pm 0.9}$}&{\large $47.6_{\pm 0.2}$}\\
&\cellcolor{gray!30}\underline{{\large $\textbf{74.4}_{\pm 1.0}$}}&\cellcolor{gray!30}\underline{{\large $\textbf{66.5}_{\pm 0.3}$}}&\cellcolor{gray!30}\underline{{\large $\textbf{80.4}_{\pm 0.5}$}}&\cellcolor{gray!30}\underline{{\large $\textbf{78.9}_{\pm 0.1}$}}&\cellcolor{gray!30}\underline{{\large $\textbf{84.7}_{\pm 0.3}$}}&\cellcolor{gray!30}\underline{{\large $\textbf{75.6}_{\pm 0.5}$}}&\cellcolor{gray!30}\underline{{\large $\textbf{65.9}_{\pm 0.6}$}}&\cellcolor{gray!30}\underline{{\large $\textbf{80.3}_{\pm 0.8}$}}&\cellcolor{gray!30}\underline{{\large $\textbf{83.7}_{\pm 0.5}$}}&\cellcolor{gray!30}\underline{{\large $\textbf{82.9}_{\pm 0.3}$}}&\cellcolor{gray!30}\underline{{\large $\textbf{74.9}_{\pm 0.9}$}}&\cellcolor{gray!30}\underline{{\large $\textbf{66.7}_{\pm 0.7}$}}&\cellcolor{gray!30}\underline{{\large $\textbf{79.7}_{\pm 0.2}$}}&\cellcolor{gray!30}\underline{{\large $\textbf{81.9}_{\pm 0.3}$}}&\cellcolor{gray!30}\underline{{\large $\textbf{82.9}_{\pm 0.1}$}}\\
\midrule
\multirow{4}{*}{8}&{{\large ${52.1}_{\pm 3.0}$}}&{{\large ${60.4}_{\pm 0.8}$}}&{{\large ${37.4}_{\pm 0.9}$}}&{{\large ${39.3}_{\pm 0.5}$}}&{{\large ${43.4}_{\pm 1.4}$}}&{\large $-$}&{\large $-$}&{\large $-$}&{\large $-$}&{\large $-$}&{\large $-$}&{\large $-$}&{\large $-$}&{\large $-$}&{\large $-$}\\
&{\large $34.1_{\pm 1.2}$}&{\large $56.7_{\pm 0.9}$}&{\large $36.1_{\pm 0.9}$}&{\large $33.3_{\pm 0.3}$}&{\large $29.1_{\pm 0.4}$}&{\large $-$}&{\large $-$}&{\large $-$}&{\large $-$}&{\large $-$}&{\large $-$}&{\large $-$}&{\large $-$}&{\large $-$}&{\large $-$}\\
&{\large $56.6_{\pm 0.4}$}&{\large $52.8_{\pm 0.2}$}&{\large $41.0_{\pm 0.6}$}&{\large $40.1_{\pm 0.3}$}&{\large $61.8_{\pm 0.6}$}&{\large $65.4_{\pm 0.4}$}&{\large $46.0_{\pm 0.9}$}&{\large $23.7_{\pm 0.9}$}&{\large $39.0_{\pm 0.4}$}&{\large $54.2_{\pm 0.4}$}&{\large $49.3_{\pm 0.3}$}&{\large $55.5_{\pm 0.3}$}&{\large $30.6_{\pm 0.9}$}&{\large $41.0_{\pm 0.4}$}&{\large $61.9_{\pm 0.5}$}\\
&\cellcolor{gray!30}\underline{{\large $\textbf{74.9}_{\pm 1.2}$}}&\cellcolor{gray!30}\underline{{\large $\textbf{66.6}_{\pm 0.6}$}}&\cellcolor{gray!30}\underline{{\large $\textbf{82.3}_{\pm 0.0}$}}&\cellcolor{gray!30}\underline{{\large $\textbf{79.6}_{\pm 0.1}$}}&\cellcolor{gray!30}\underline{{\large $\textbf{85.8}_{\pm 0.3}$}}&\cellcolor{gray!30}\underline{{\large $\textbf{76.6}_{\pm 0.4}$}}&\cellcolor{gray!30}\underline{{\large $\textbf{65.9}_{\pm 0.6}$}}&\cellcolor{gray!30}\underline{{\large $\textbf{82.5}_{\pm 1.6}$}}&\cellcolor{gray!30}\underline{{\large $\textbf{84.8}_{\pm 0.3}$}}&\cellcolor{gray!30}\underline{{\large $\textbf{84.3}_{\pm 0.3}$}}&\cellcolor{gray!30}\underline{{\large $\textbf{75.5}_{\pm 0.6}$}}&\cellcolor{gray!30}\underline{{\large $\textbf{67.0}_{\pm 0.2}$}}&\cellcolor{gray!30}\underline{{\large $\textbf{82.3}_{\pm 0.4}$}}&\cellcolor{gray!30}\underline{{\large $\textbf{82.8}_{\pm 0.0}$}}&\cellcolor{gray!30}\underline{{\large $\textbf{84.1}_{\pm 0.2}$}}\\
\midrule
\multirow{4}{*}{16}&{{\large ${52.9}_{\pm 2.7}$}}&{{\large ${60.4}_{\pm 0.8}$}}&{{\large ${37.4}_{\pm 0.9}$}}&{{\large ${39.3}_{\pm 0.5}$}}&{{\large ${44.0}_{\pm 1.4}$}}&{\large $-$}&{\large $-$}&{\large $-$}&{\large $-$}&{\large $-$}&{\large $-$}&{\large $-$}&{\large $-$}&{\large $-$}&{\large $-$}\\
&{\large $32.6_{\pm 0.1}$}&{\large $56.9_{\pm 0.9}$}&{\large $34.6_{\pm 1.2}$}&{\large $31.4_{\pm 0.4}$}&{\large $34.8_{\pm 1.2}$}&{\large $-$}&{\large $-$}&{\large $-$}&{\large $-$}&{\large $-$}&{\large $-$}&{\large $-$}&{\large $-$}&{\large $-$}&{\large $-$}\\
&{\large $68.8_{\pm 0.1}$}&{\large $64.8_{\pm 0.2}$}&{\large $32.6_{\pm 0.9}$}&{\large $46.7_{\pm 0.9}$}&{\large $67.8_{\pm 0.3}$}&{\large $70.6_{\pm 0.9}$}&{\large $56.3_{\pm 0.9}$}&{\large $35.6_{\pm 0.9}$}&{\large $46.8_{\pm 0.9}$}&{\large $58.6_{\pm 0.9}$}&{\large $66.5_{\pm 0.1}$}&{\large $63.9_{\pm 0.4}$}&{\large $39.1_{\pm 0.3}$}&{\large $38.8_{\pm 0.9}$}&{\large $69.4_{\pm 0.9}$}\\
&\cellcolor{gray!30}\underline{{\large $\textbf{75.1}_{\pm 1.1}$}}&\cellcolor{gray!30}\underline{{\large $\textbf{66.6}_{\pm 0.6}$}}&\cellcolor{gray!30}\underline{{\large $\textbf{83.1}_{\pm 0.4}$}}&\cellcolor{gray!30}\underline{{\large $\textbf{79.8}_{\pm 0.1}$}}&\cellcolor{gray!30}\underline{{\large $\textbf{86.6}_{\pm 0.3}$}}&\cellcolor{gray!30}\underline{{\large $\textbf{76.8}_{\pm 0.8}$}}&\cellcolor{gray!30}\underline{{\large $\textbf{66.1}_{\pm 0.3}$}}&\cellcolor{gray!30}\underline{{\large $\textbf{83.8}_{\pm 1.1}$}}&\cellcolor{gray!30}\underline{{\large $\textbf{85.5}_{\pm 0.2}$}}&\cellcolor{gray!30}\underline{{\large $\textbf{85.3}_{\pm 0.3}$}}&\cellcolor{gray!30}\underline{{\large $\textbf{75.7}_{\pm 0.9}$}}&\cellcolor{gray!30}\underline{{\large $\textbf{66.8}_{\pm 0.1}$}}&\cellcolor{gray!30}\underline{{\large $\textbf{83.2}_{\pm 0.4}$}}&\cellcolor{gray!30}\underline{{\large $\textbf{83.3}_{\pm 0.3}$}}&\cellcolor{gray!30}\underline{{\large $\textbf{85.2}_{\pm 0.1}$}}\\

\bottomrule
\end{tabular}
    }

\end{subtable}

\caption{Classification accuracy. The setup is identical to Table \ref{tab:compare_with_GAP_and_deva}, and the graph setting is inductive.}
\label{tab:compare_with_GAP_and_deva_inductive}

\end{table*}

\subsection{Proof of Theorem \ref{thm:privacy_guarantee}}\label{app:proof_privacy_guarantee}

\begin{proof}
    The $T$ multiplication exists because $T$-fold adaptive composition results in linear add-up in R\'enyi divergence. It is sufficient to only upper-bound the divergence 
    for one iteration. 
    
    1) Suppose we are in this case: differing node $z$ is sampled, such as the case in Figure \ref{fig:illustration_with_z}, the final sub-graph container \textbf{$G'$ has only 1 more sub-graph than  $G^*$}, which centers around $z$. Denote $\Bar{g}^t|_{G^*}$ as the output shown at line \ref{alg:nodedp_non_private_gradient} of Algorithm \ref{alg:nodedp_training} if the graph container is $G^*$ due to input graph $\mathcal{G}^*$, and similar, $\Bar{g}^t|_{G'}$ due to input graph $\mathcal{G}'$, we have $$\max\|\Bar{g}^t|_{G^*} - \Bar{g}^t|_{G'} \|_2 = \max\|\hat{g}_z\|_2 = 0.5$$ where $\hat{g}_z$ is the clipped gradient computed on sub-graph centering around $z$. Conditioned on this case, denote the distribution for $\mathcal{G}^*$ and $\mathcal{G}'$ at line \ref{alg:nodedp_private_gradient} as $g^t|_{G^*,0.5}$ and $g^t|_{G',0.5}$, respectively. We have $\mathcal{D}_{\alpha}(g^t|_{G^*,0.5}||g^t|_{G',0.5})\leq\mathcal{D}_{\alpha}(\mathcal{N}(0,\sigma^2)||\mathcal{N}(1/2,\sigma^2))$.
    This case happens with probability $q_b$ as $z$ is sampled with probability $q_b$.

    2) Suppose we are in another case: differing node $z$ is \underline{\textbf{not}} sampled and $k$ out-pointing nodes of $z$ are sampled, and all of those $k$ nodes also sample $z$ as neighbors, as depicted in Figure \ref{fig:illustration_without_z} where we draw the scenario when $k=1$. Note that container  \textbf{$G^*, G'$ differ in $k$ sub-graphs}, then, we have $$\max\|\Bar{g}^t|_{G^*} - \Bar{g}^t|_{G'} \|_2 = 2\times k \times 0.5=k.$$ Conditioned on this case and in a similar way, denote the distribution for $\mathcal{G}^*$ and $\mathcal{G}'$ at line \ref{alg:nodedp_private_gradient} as $g^t|_{G^*,k}$ and $g^t|_{G',k}$, respectively. We have $\mathcal{D}_{\alpha}(g^t|_{G^*,k}||g^t|_{G',k})\leq\mathcal{D}_{\alpha}(\mathcal{N}(0,\sigma^2)||\mathcal{N}(k,\sigma^2))$ . This case happen with probability $(1-q_b)\mathbf{Bi}(k; D_{ot}, \frac{q_bM}{D_{ot}})$, where $D_{ot}$ is out-degree of $z$.

    And by corollary \ref{cor:vmog_to_mog}, now it it clear that the privacy of algorithm \ref{alg:nodedp_training} fits into MoG defined in Definition \ref{def:moG}. 

    Note that we ensure a guarantee for any node with any $D_{ot}$, hence taking maximum over all possible $D_{ot}$.
    A final remark is: to have valid privacy bound for a randomized algorithm $\mathcal{M}$ (denote $\mathcal{M}^*=\mathcal{M}(\mathcal{G^*})$ and $\mathcal{M}'=\mathcal{M}(\mathcal{G'})$ in our case), due to symmetric of adjacency, we must  take the maximal of $\mathcal{D}_{\alpha}(P||Q)$ and $\mathcal{D}_{\alpha}(Q||P)$.  
\end{proof}


\subsection{Proof of Theorem \ref{thm:private_embedding_impossible_special}}\label{appendix:proof_private_node_embedding_special}

\begin{proof}
we uniformly randomly pick node $i$ and w.l.o.g., suppose the class of interest is $1$. The goal is to upper-bound the precision w.r.t. label $1$, {\it i.e.}, $\operatorname{Pr}[\mathbf{Y}[i]=1|h(u_i)=1]$. By Bayes' theorem, we have:
\begin{equation}\nonumber
\begin{aligned}
    \operatorname{Pr}[\mathbf{Y}[i]=1|h(u_i)=1] = \frac{\operatorname{Pr}[h(u_i)=1|\mathbf{Y}[i]=1] \operatorname{Pr}[\mathbf{Y}[i]=1]}{\operatorname{Pr}[h(u_i)=1]}
\end{aligned}
\end{equation}
Note that the classifier is Class-1-Aligned,
{\it i.e.}, $\operatorname{Pr}[\mathbf{Y}[i]=1]=\operatorname{Pr}[h(u_i)=1]$. Hence, we have:
\begin{equation}\label{equ:presion_equ}
\begin{aligned}
    \operatorname{Pr}[\mathbf{Y}[i]=1|h(u_i)=1] = \operatorname{Pr}[h(u_i)=1|\mathbf{Y}[i]=1]
\end{aligned}
\end{equation}
Based on Definition \ref{def:private_embedding}, we know Equation \eqref{equ:private_node_embedding} holds for any adjacent graph pair $(\mathcal{G}^*, \mathcal{G'})$, and this also includes the case that the current node $i$ of interest is the differing node. This means that the class of node with ID $i$ in $\mathcal{G}^*$ and $\mathcal{G'}$ can differ, {\it i.e.}, in $\mathcal{G}^*$ we have $\mathbf{Y}[i]=1$ and in $\mathcal{G'}$ we can have $\mathbf{Y}[i]=2$. Define $S = \{u_i|h(u_i)=1\}$, we then have:
    \begin{equation}\nonumber
    \begin{aligned}
        \operatorname{Pr}[h(u_i)=1|\mathbf{Y}[i]=1] \leq e^{\varepsilon} \operatorname{Pr}[h(u_i)=1|\mathbf{Y}[i]=2] + \delta,
    \end{aligned}
    \end{equation}
    similarly, we also have:
    \begin{equation}\nonumber
    \begin{aligned}
        \operatorname{Pr}[h(u_i)=1|\mathbf{Y}[i]=1] \leq e^{\varepsilon} \operatorname{Pr}[h(u_i)=1|\mathbf{Y}[i]=3] + \delta,
    \end{aligned}
    \end{equation}
    \begin{equation}\nonumber
        \cdots
    \end{equation}
    \begin{equation}\nonumber
    \begin{aligned}
        \operatorname{Pr}[h(u_i)=1|\mathbf{Y}[i]=1] \leq e^{\varepsilon} \operatorname{Pr}[h(u_i)=1|\mathbf{Y}[i]=C] + \delta,
    \end{aligned}
    \end{equation}
    note that the events $\mathbf{Y}[i]=l, \forall l\in {1,2,\cdots, C}$ are mutually exclusive, then we have:
    \begin{equation}\nonumber
    \begin{aligned}
         \operatorname{Pr}[h(u_i)=1|\mathbf{Y}[i]=1] = 1-\sum_{l=2}^{C}\operatorname{Pr}[h(u_i)=1|\mathbf{Y}[i]=l],
    \end{aligned}
    \end{equation}
    adding both the right-hand side and left-hand side of the above $C-1$ inequalities, rearrange, we get:
    \begin{equation}\nonumber
    \begin{aligned}
        \operatorname{Pr}[h(u_i)=1|\mathbf{Y}[i]=1] \leq \frac{e^{\varepsilon}+\delta(C-1)}{C-1+e^{\varepsilon}}
    \end{aligned}
    \end{equation}
    based on Equation \eqref{equ:presion_equ}, we have the final result:
    \begin{equation}\nonumber
    \begin{aligned}
        \operatorname{Pr}[\mathbf{Y}[i]=1|h(u_i)=1] \leq \frac{e^{\varepsilon}+\delta(C-1)}{C-1+e^{\varepsilon}}
    \end{aligned}
    \end{equation}

\end{proof}

\begin{figure}[!ht] 
    \centering
    \subfloat[$\varepsilon=2$]
    {\includegraphics[width=.99\linewidth]{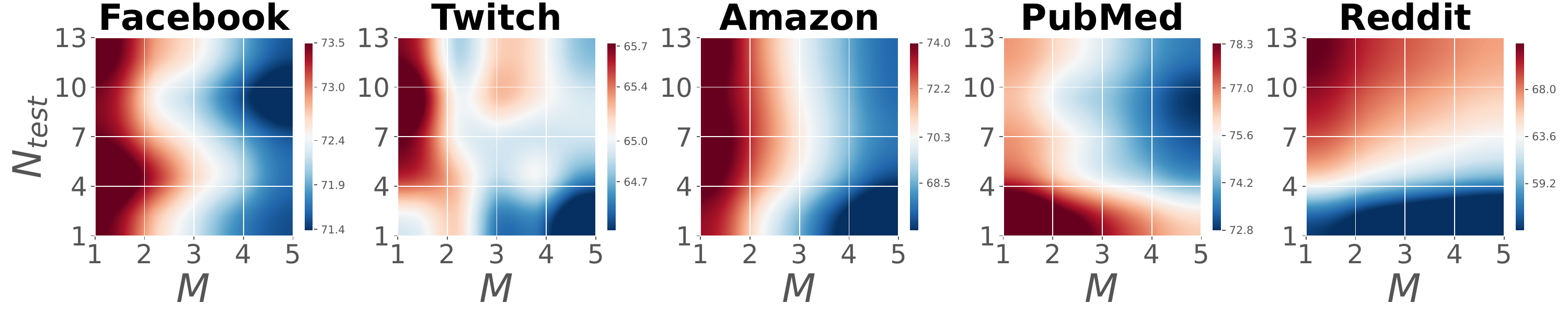}
    \label{fig:trade_offs_ns_gcn_eps1_inducitve}}\\
    \subfloat[$\varepsilon=\infty$]{\includegraphics[width=.99\linewidth]{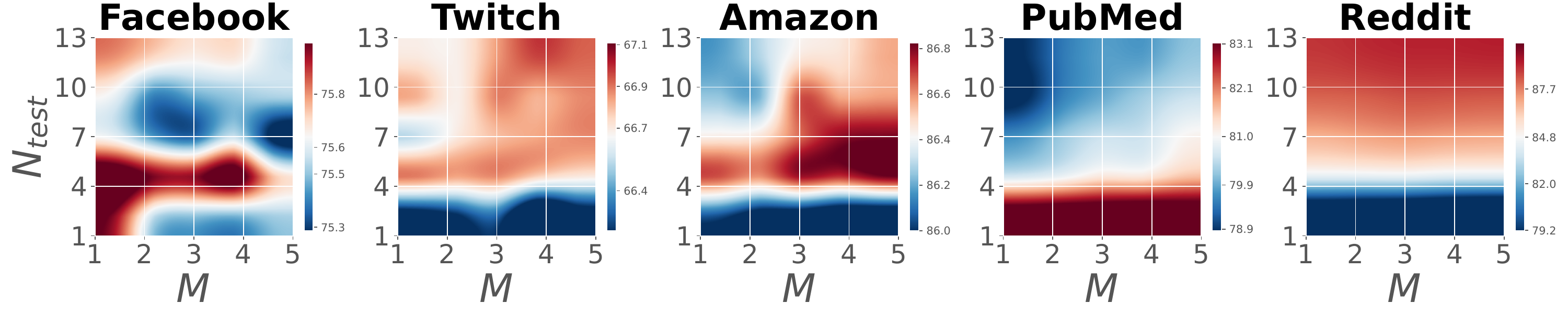}
    \label{fig:trade_offs_ns_gcn_eps10000_inducitve}}\\
    
    \caption{Trade-offs in neighborhood sampling in inductive graph setting. The setup is identical to Figure \ref{fig:trade_offs_ns_gcn}}
    \label{fig:trade_offs_ns_gcn_inducitve} 
    
\end{figure}

\begin{figure}[!ht] 

    \centering
    \subfloat[$\varepsilon=2$]
    {\includegraphics[width=.49\linewidth]{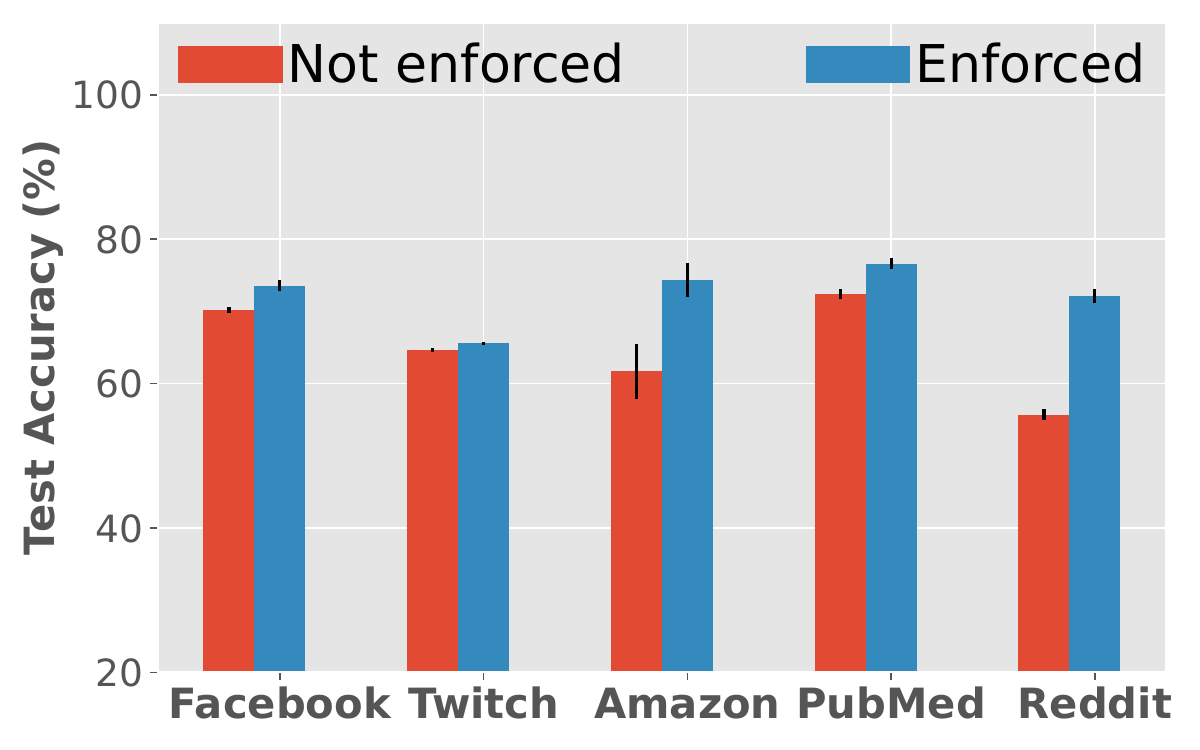}
    \label{fig:no_overlap_2_inductive}}
    \subfloat[$\varepsilon=16$]{\includegraphics[width=.49\linewidth]{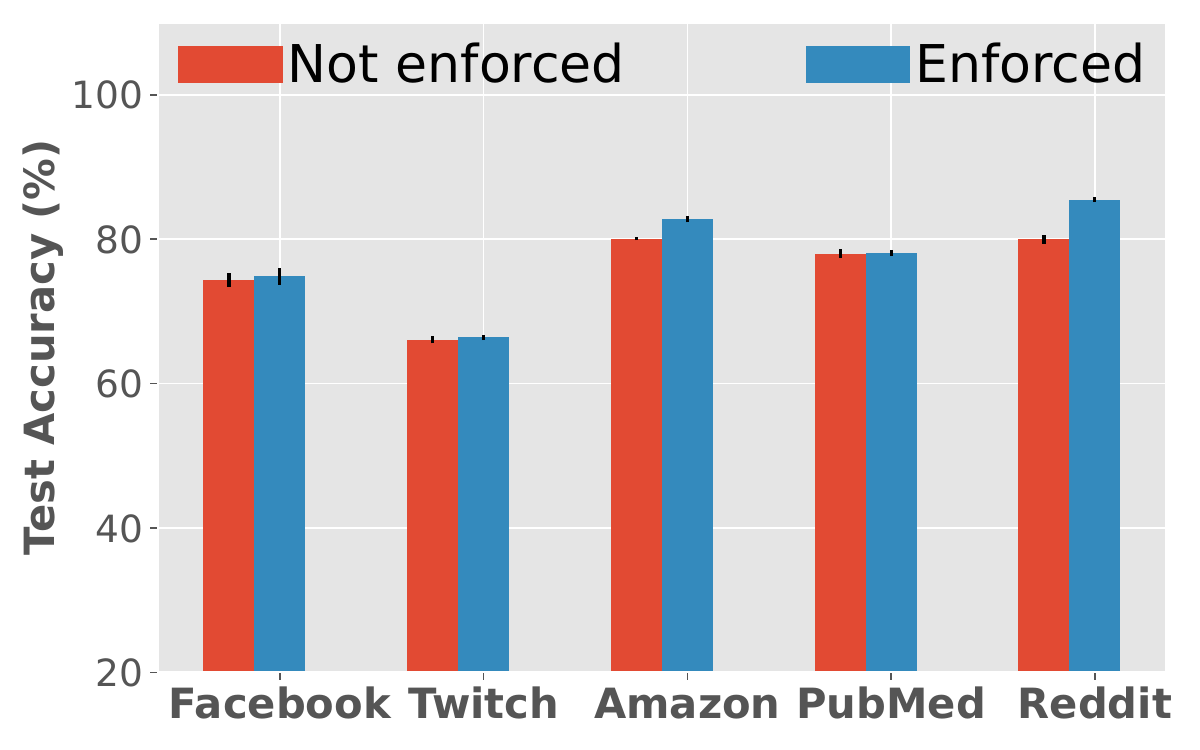}
    \label{fig:no_overlap_16_inductive}}\\

    \caption{Performance comparison. The setup is identical to Figure \ref{fig:no_overlap}, and the graph setting is inductive.}
    \label{fig:no_overlap_inductive} 
    
\end{figure}

\section{More experimental results}\label{app:add_exp}

Additional counterpart results on the inductive graph setting are in this section. The accuracy is shown in Table \ref{tab:compare_with_GAP_and_deva_inductive}, and the investigation for the inductive graph is presented in Figure \ref{fig:no_overlap_inductive}. Ablation study on neighborhood sampling under is provided in Figure \ref{fig:trade_offs_ns_gcn_inducitve}.

\end{document}